\documentclass[,referee,sn-basic]{sn-jnl}%

\usepackage{booktabs}
\usepackage{multirow}
\usepackage[figuresright]{rotating}
\usepackage{graphicx}%
\usepackage{multirow}%
\usepackage{amsmath,amssymb,amsfonts}%
\usepackage{amsthm}%
\usepackage{mathrsfs}%
\usepackage[title]{appendix}%
\usepackage[table]{xcolor}%
\usepackage{textcomp}%
\usepackage{manyfoot}%
\usepackage{booktabs}%
\usepackage{algorithm}%
\usepackage{algorithmicx}%
\usepackage{algpseudocode}%
\usepackage{listings}%
\usepackage{comment}

\theoremstyle{thmstyleone}%

\theoremstyle{thmstyletwo}%
\newtheorem{remark}{Remark}%

\theoremstyle{thmstylethree}%
\newtheorem{definition}{Definition}%

\definecolor{aqcolor}{RGB}{0,0,255}
\definecolor{lwcolor}{RGB}{0,0,0}
\definecolor{binbincolor}{RGB}{0,0,0}
\definecolor{hinacolor}{RGB}{138,43,226}

\definecolor{anqicolor}{RGB}{0,0,0}
\definecolor{bincolor}{RGB}{0,0,0}
\definecolor{modify}{RGB}{0,0,0}

\newcommand{\binwang}[1]{\textcolor{bincolor}{#1}}
\newcommand{\bwang}[1]{\textcolor{binbincolor}{#1}}

\newcommand{\aq}[1]{\textcolor{aqcolor}{#1}}
\newcommand{\aqthenbw}[1]{\textcolor{lwcolor}{#1}}

\definecolor{change}{RGB}{180,150,0}

\begin{document}



\title[Article Title]{Unlocking air traffic flow prediction through microscopic aircraft-state modeling}




\author[1]{\fnm{Bin} \sur{Wang}}\email{wangbin9545@ouc.edu.cn}
\equalcont{These authors contributed equally to this work.}

\author[1]{\fnm{Anqi} \sur{Liu}}
\equalcont{These co-first authors contributed equally to this work.}\email{liuanqi8378@stu.ouc.edu.cn}
\author[2]{\fnm{Jiangtao} \sur{Zhao}}
\author[3]{\fnm{Yanyong} \sur{Huang}}
\author[1]{\fnm{Hina} \sur{Birahmani}}

\author[1]{\fnm{Peilan} \sur{He}}
\author[1]{\fnm{Guiyuan} \sur{Jiang}}

\author*[1]{\fnm{Feng} \sur{Hong}}
\author[1]{\fnm{Yanwei} \sur{Yu}}
\author[4]{\fnm{Yuanyuan} \sur{Hou}}


\author[5]{\fnm{Tianrui} \sur{Li}}

\affil[1]{\orgdiv{Faculty of Information Science and Engineering}, \orgname{Ocean University of China}, \orgaddress{\city{Qingdao}, \postcode{266100}, \state{Shandong}, \country{China}}}

\affil[2]{\orgdiv{Sanya Oceanographic Institution}, \orgname{Ocean University of China}, \orgaddress{\city{Sanya}, \postcode{572022}, \state{Hainan}, \country{China}}}

\affil[3]{\orgdiv{Joint Laboratory of Data Science and Business Intelligence, Institute of Statistical Interdisciplinary Research,} \orgname{Southwestern University of Finance and Economics}, \orgaddress{\city{Chengdu}, \postcode{611130}, \state{Sichuan}, \country{China}}}

\affil[4]{\orgdiv{Department of Rehabilitation Medicine}, \orgname{The Affiliated Hospital of Qingdao University}, \orgaddress{\city{Qingdao}, \postcode{266100}, \state{Shandong}, \country{China}}}

\affil[5]{\orgdiv{School of Computing and Artificial Intelligence}, \orgname{Southwest Jiaotong University}, \orgaddress{\city{Chengdu}, \postcode{610000}, \state{Sichuan}, \country{China}}}

\abstract{Short-term air traffic flow prediction in terminal airspace is essential for proactive air traffic management. Existing approaches predominantly model traffic flow as aggregated time series. {However, traffic dynamics are governed by aircraft states and their interactions in continuous airspace.} Such aggregation obscures fine-grained information{,} including aircraft kinematics, boundary interactions, and control-intent cues. Here we present AeroSense, a state-to-flow modeling {paradigm} that predicts future traffic flow directly from instantaneous airspace situations represented as dynamic sets of aircraft states derived from ADS-B trajectories. By establishing an end-to-end mapping from microscopic aircraft states to future regional traffic flow, AeroSense preserves aircraft-level dynamics while naturally accommodating varying traffic density, and avoids reliance on historical look-back windows. Experiments on a large-scale real-world dataset show that AeroSense exhibits strong predictive accuracy and robustness compared with time series-based forecasting approaches, \bwang{without requiring exhaustive hyperparameter tuning}. These findings suggest that aircraft-state situation modeling provides a promising alternative to conventional time-series forecasting in air traffic flow management.}


\maketitle

\section{Introduction}\label{sec1}

Air traffic management (ATM) is shifting toward predictive, trajectory-informed operations, where future system states are anticipated rather than reactively controlled \citep{shi2021ConstrainedLSTM, Guo2024FlightBERTpp}. This transition is reflected in major modernization programmes, including the Single European Sky ATM Research (SESAR) initiative in Europe and the Next Generation Air Transportation System (NextGen) in the United States \citep{brooker2008sesar}, both of which promote the use of  trajectory information, such as automatic dependent surveillance-broadcast (ADS-B) data \citep{zeng2022aircraft}, to support coordinated decision-making across the airspace \citep{nagaoka2014review}. In this context, understanding the evolution of short-term traffic flow is essential in air traffic flow management (ATFM), especially for managing airspace congestion, balancing sector demand and capacity, and ensuring operational safety \citep{lin2019deep}. Accordingly, air traffic flow prediction  is not merely a forecasting task, but a core capability for enabling proactive and resilient ATFM \citep{yan2023multi, wandelt2025flight}. {These challenges become especially pronounced in operationally complex airspaces. }

\begin{figure}[!ht]
  \centering
  \includegraphics[width=\linewidth]{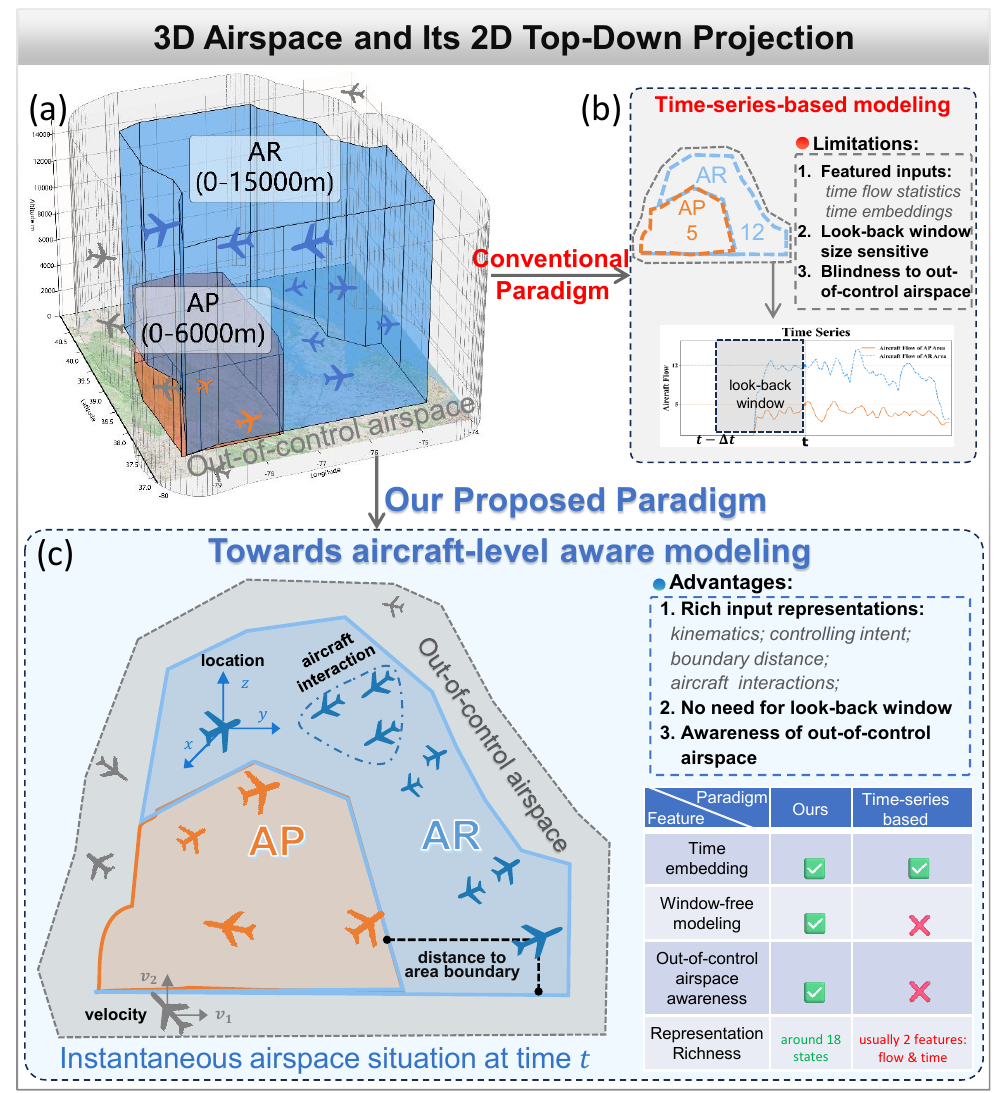} 
  \caption{\textbf{The motivation of aircraft-level state modeling.} The proposed aircraft-level modeling paradigm directly models airspace situation for flow prediction of $(t+\Delta t]$ using only instantaneous aircraft-level states at time $t$, without requiring the look-back window.
}
  \label{fig:motivation}
\end{figure}

Among different airspace types, terminal airspace  (TA) is particularly critical \citep{jurinic2024defining,li2024airspace,yin2025aircraft}. It serves the interface between en-route airspace and airport surface operations, where arrivals and departures are densely concentrated and frequently subject to air traffic control (ATC) interventions. In such highly dynamic environments, even small fluctuations in traffic conditions can escalate into system-level disruptions \citep{chen2016network, jiang2024characteristics}. The TA is typically divided into two key functional regions \footnote{{The \underline{A}irspace \underline{C}ontrol \underline{R}egion is interchangeably abbreviated as AC. In this study, however, we adopt the abbreviation AR following the terminology accustomed to the local ATC authority. The AP corresponds to low-altitude airspace near the airport surface (typically 0--6000m), while the AR covers higher-altitude airspace (typically up to 15,000m) with substantially denser traffic flows. These regions represent the primary focus of ATC operations, whereas observable traffic outside AP and AR lies beyond the area control of the local ATC authority.}}: the {\underline{Ap}proach Airspace} (AP) and the {\underline{A}irspace  Control \underline{R}egion} (AR), as illustrated in Fig. ~\ref{fig:motivation}(a). The AP corresponds to low-altitude airspace near the airport surface and generally exhibits lower traffic volumes, whereas the AR covers higher-altitude airspace with significantly denser  flows. Accurate short-term air traffic flow prediction in both regions is therefore crucial for proactive traffic management \citep{li2024airspace}, enabling controllers to implement sequencing, holding, and spacing strategies ahead of demand surges \citep{jurinic2024defining}. 

\bwang{Aircraft entry counts over a defined airspace and time interval are established ATM measures for characterizing traffic demand and supporting demand--capacity assessment. ICAO expresses declared capacity in terms of the number of aircraft entering a specified portion of airspace within a given period ~\citep{icao2018annex11}, while EUROCONTROL uses entry counts together with occupancy counts for traffic monitoring and demand--capacity management ~\citep{dalichampt2007hourly,eurocontrol2024flowconops}. Accordingly, we use short-term aircraft counts in AP and AR as an interpretable measure of
regional traffic demand.} \bwang{Although flight plans, estimated entry times, and coordination information provide important prior knowledge, they do not uniquely determine the traffic demand that will materialize within a fixed short-term interval. Active trajectories are continuously updated using surveillance observations and controller inputs, and aircraft may still deviate from their latest planned sector entry within a 15-minute horizon~\citep{icao2025atmas,klotergens2023predicting}. This plan--execution gap motivates data-driven forecasting of realized regional demand from the current traffic state. Such surveillance-based forecasting is intended to complement, rather than replace, flight-plan-based systems. Moreover, the 15-minute horizon corresponds to an established tactical planning timescale in ATM~\citep{sesar2023solution53b}.}

%

With the widespread availability of high-resolution ADS-B trajectory data \citep{strohmeier2014realities, patrikar2025image}, data-driven approaches have become the dominant methodology for air traffic flow modeling \citep{du2024spatial}. 
\aqthenbw{Prior aircraft trajectory clustering studies have shown that characteristic air traffic patterns can provide practical value for terminal airspace operations.  \citep{olive2019trajectory}.} 
Early studies primarily employed statistical learning models, such as support vector machines and gradient boosting trees, to forecast traffic volume from historical observations \citep{rebollo2014characterization}. More recently, {deep learning approaches \citep{yan2022deep, yan2023multi}, particularly graph neural networks \citep{zhang2025short}} and Transformer-based forecasting models \citep{ma2024text}, have shown strong capability in capturing spatial interactions and temporal dynamics in air traffic flow \citep{wu2024long}.

\aqthenbw{Aircraft-level representations have been explored in several ATM prediction tasks. For example, \citet{pang2023workload} represented evolving traffic configurations as dynamic aircraft graphs for controller workload prediction, while \citet{vos2024transformer} predicted individual aircraft trajectories and subsequently derived sector demand from their predicted airspace crossings. Multi-aircraft interactions have also been considered in arrival time prediction through attention-based models \citep{nguyen2025multi} and probabilistic formulations that model both individual trajectory patterns and inter-aircraft dependencies \citep{kim2026probabilistic}. These studies demonstrate the usefulness of aircraft-level information across several ATM prediction tasks.}

Despite these advances,  \aqthenbw{regional air traffic flow forecasting is still predominantly formulated as }a \textit{macroscopic time-series forecasting paradigm}. Aircraft trajectories are first aggregated into macroscopic flow sequences or spatio-temporal tensors \citep{lin2019deep, yan2023multi, ma2024text}, after which prediction is performed on these derived {time-series-based flow} representations (e.g., aggregate traffic counts such as 5 aircraft in AP and 12 aircraft in AR, as illustrated in Fig.~\ref{fig:motivation}(b)). {For example, \cite{lin2019deep} proposed to project the airspace onto a horizontal view and partitioned the resulting view into fixed-size spatial grids. The number of aircraft within each grid cell was then aggregated as local traffic flow, producing an image-like representation in which each pixel encodes the aircraft count of the corresponding region. This formulation allows traffic flow prediction to be cast as an image modeling problem and subsequently addressed using CNN. Such methods primarily rely on statistical correlations learned from aggregated spatio-temporal flow, occasionally augmented with conditions such as temporal embeddings  \citep{niu2026phaseformer} and weather factors \citep{zhang2023flight}. Although operationally convenient and intuitive}, this conventional paradigm largely overlooks microscopic aircraft-level dynamics and interactions. Compressing aircraft states into predefined time-series representations inevitably introduces discretization artifacts, sparsity, and information loss, particularly for fine-grained kinematic cues such as velocity direction and boundary proximity. Consistent with this concern, trajectory-based analyses have shown that aircraft trajectories can reveal significant operational events associated with weather disruptions, traffic sequencing, and tactical air traffic control interventions \citep{nguyen2025multi, olive2020detection}. This evidence suggests that aircraft-level trajectories retain decision-relevant information that may be obscured by macroscopic flow aggregation. Therefore, a key limitation of existing approaches may lie not only in their forecasting architectures, but also in the underlying representation paradigm used to characterize air traffic dynamics.

We argue that, at any given moment $t$,  the airspace situation in the TA is more naturally characterized not as a time series, but as a \textit{dynamic, unordered set of aircraft}, whose cardinality continuously changes due to aircraft entry and exit events. This perspective raises a fundamental requirement: \textit{Can instantaneous microscopic aircraft states provide sufficient information for accurate future air traffic flow prediction without relying on historical flow sequences?}

\begin{figure*}[!htb]
  \centering
  \includegraphics[width=\linewidth]{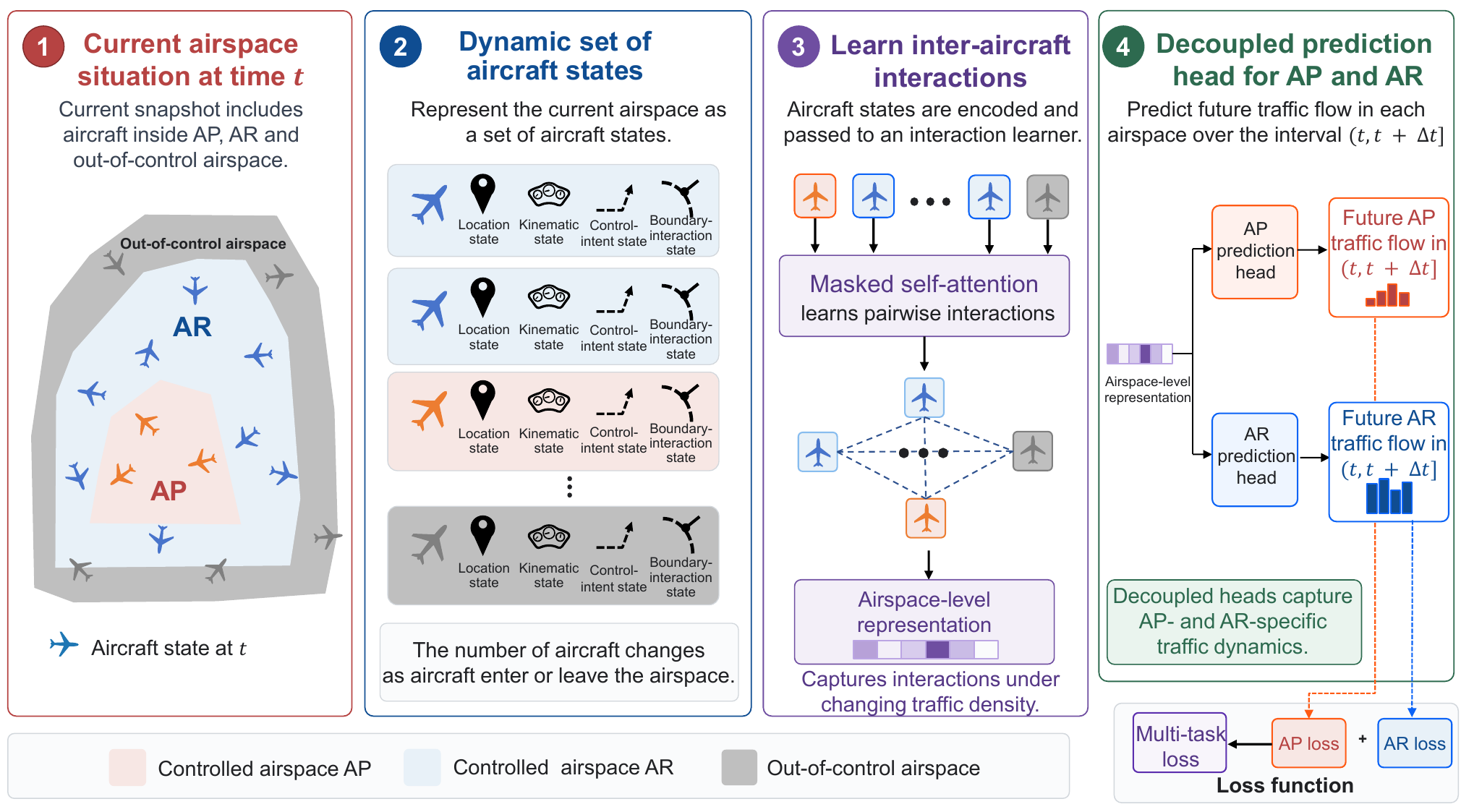} 
  \vspace{-1em}
  \caption{\textbf{Overview of the proposed AeroSense framework.} {
  (1) The instantaneous airspace situation at any time $t$ is constructed from real-time streaming ADS-B observations without requiring historical trajectory storage or long look-back windows. 
  (2) The current airspace is represented as a dynamic, variable-cardinality set of aircraft states, where each aircraft encodes kinematic information, control-intent cues, boundary interactions, and temporal contextual cues. 
  (3) These aircraft-level states are processed by masked self-attention to capture inter-aircraft interactions under varying traffic density and are subsequently aggregated into an airspace-level representation. 
  (4) Two decoupled prediction heads are employed to separately estimate future traffic flows in the AP and AR regions over the prediction horizon $(t, t+\Delta t]$, with both branches jointly optimized during training.}
}\label{fig:overview}
\end{figure*}

Motivated by this question, we propose \textit{AeroSense}, an \textit{\underline{aero}nautical} aircraft-level state-to-flow modeling framework that \underline{\textit{senses}} future traffic flow directly from microscopic aircraft states using ADS-B trajectory data, as illustrated in Fig.~\ref{fig:overview}. Instead of transforming aircraft trajectories into aggregated macroscopic statistics prior to prediction, AeroSense formulates the instantaneous airspace situation at time $t$ as a dynamic, variable-cardinality set of aircraft states and establishes an end-to-end mapping from this set to future airspace-level flows at horizon $t+\Delta t$. This formulation preserves fine-grained aircraft kinematics, naturally accommodates varying traffic density, and eliminates the dependence on manually specified \textit{look-back window} hyperparameters that are central to conventional time-series forecasting pipelines. Moreover, because AeroSense operates directly on instantaneous airspace situations rather than long historical sequences, it is inherently suitable for real-time streaming inference without requiring persistent historical ADS-B trajectory storage. \bwang{From an online inference perspective, this distinction is advantageous because historical look-back sequences are no longer mandatory inputs for each forecast. A conventional time-series model predicts from
a fixed-length sequence:
$
\hat{\mathbf{Y}}_{t+1}
= F\left(\mathbf{Y}_{t-L+1},\ldots,\mathbf{Y}_{t}
\right)$,
where the look-back length $L$ is itself an important hyperparameter
that typically requires exhaustive tuning for optimal performance. Moreover, The corresponding flow buffer must be
continuously aggregated, maintained, and synchronized before inference. If any $\mathbf{Y}_{i}$ is delayed, missing, or reconstructed under an inconsistent aggregation boundary, the input sequence must first be repaired or reconstructed. In contrast, AeroSense predicts directly from the latest aircraft-state
set $\mathcal{S}_{t}$: $
\hat{\mathbf{Y}}_{t}
=
F(\mathcal{S}_{t}),
$
allowing each forecast to be generated from the current airspace situation without  maintaining the preceding $L$-step historical flow window.}

To realize this formulation, AeroSense incorporates a situation-aware aircraft state representation that encodes physically meaningful cues, including GPS location, boundary interactions, control-intent cues, etc., into the airspace situation modeling process. 
Building upon these representations, we further develop a permutation-invariant architecture that combines variable-cardinality set handling, masked self-attention for inter-aircraft interaction modeling, and task-decoupled prediction heads to capture the heterogeneous flow dynamics of the AP and AR airspaces.

Our contributions are summarized as follows:
\begin{enumerate}
    \item We \aqthenbw{reformulate} air traffic flow prediction from a macroscopic time-series forecasting problem into a \emph{state-to-flow} learning problem. Instead of relying on aggregated historical flow sequences, we directly model the instantaneous airspace situation as a variable-cardinality set of microscopic aircraft states, better aligning the learning representation with the underlying physical state of air traffic.

    \item We propose AeroSense, an aircraft-level state-to-flow modeling framework that establishes an end-to-end mapping from aircraft-level states to future airspace traffic flows. The framework naturally supports dynamic traffic density, eliminates the need for manually specified look-back windows, and mitigates information loss caused by spatio-temporal aggregation.

    \item We introduce physically grounded aircraft-state representations that encode boundary interactions, partial control-intent cues, and other flight-dynamics information as inductive biases. This design enables the model to reason about traffic evolution from underlying aircraft dynamics rather than relying solely on statistical correlations.

    \item Extensive experiments on large-scale real-world terminal airspace dataset demonstrate that AeroSense, \bwang{without relying on exhaustive hyperparameter tuning, consistently outperforms representative SOTA time-series forecasting baselines and maintains strong predictive accuracy and robustness across diverse traffic conditions.}

\end{enumerate}

\section{Results}\label{sec2}

\subsection{Problem Formulation}\label{sec:problem_formulation}

\begin{definition}[Airspace Situation]\label{def:as}
The \textit{Airspace Situation} at instantaneous time $t$ is formally defined as the set  
\begin{equation}
    \mathbf{S}_t \overset{\text{def}}{=} \{ \mathbf{s}_1^{(t)}, \mathbf{s}_2^{(t)}, \dots, \mathbf{s}_{N_t}^{(t)} \}.
\end{equation}

Each element $\mathbf{s}_i^{(t)}$ is extracted from the ADS-B data and represents the state of the $i$-th aircraft, which is given by
\begin{equation}\label{ref:istate}
    \mathbf{s}_i^{(t)} = [\mathbf{p}_i, \mathbf{v}_i, \boldsymbol{\phi}_{i}]  \in \mathbb{R}^{D_{\mathrm{in}}},
\end{equation}
where $\mathbf{p}_i \in \mathbb{R}^3$ denotes the  3-D aircraft position (longitude, latitude, altitude), $\mathbf{v}_i \in \mathbb{R}^3$ denotes the 3-D velocity vector,  $\boldsymbol{\phi}_{i}$ encodes physics-related states such as boundary proximity and flight intention (detailed in Section~\ref{sec:physics_features}). Here, $D_{\mathrm{in}}$ denotes the total dimensionality of the aircraft-level state as model input vector, and $N_t = |\mathbf{S}_t|$ represents the number of aircraft present in the airspace at time $t$.  
\end{definition}
\begin{remark}\label{ref:rm1}
    Because aircraft continuously enter and leave the airspace, hence $N_t$ can vary over time, that is, $N_t \neq N_{t+1}$ in general. Therefore, $\mathbf{S}_t$ \textit{is inherently a variable-cardinality set}, which \textit{precludes the direct use of fixed-input-size  models} such as classical MLPs, CNNs or Transformers.
\end{remark}

\begin{definition}[Combined Airspace Scope]
Our prediction targets are the AP and AR, which together define the
\textit{controlled airspace scope}:
\begin{equation}
    \Omega_{\mathrm{ctr}} = \mathrm{AP} \cup \mathrm{AR}.
\end{equation}

\bwang{Let $\Omega_{\mathrm{surv}} \subset \mathbb{R}^3$ denote the surveillance domain
observable to the local ATM authority. We define the surrounding}
\textit{out-of-control airspace scope} as
\begin{equation}\label{ref:d}
    \Omega_{\mathrm{unctr}}(d)
    =
    \left\{
    p \in \Omega_{\mathrm{surv}} \setminus \Omega_{\mathrm{ctr}}
    \mid
    \mathrm{dist}(p,\partial\Omega_{\mathrm{ctr}}) \le d
    \right\},
\end{equation}
\bwang{where $\partial\Omega_{\mathrm{ctr}}$ denotes the boundary of the controlled
airspace and $d$ controls the extent of the surrounding context.}

The \textit{combined input airspace} is then
\begin{equation}
    \Omega(d)
    =
    \Omega_{\mathrm{ctr}} \cup \Omega_{\mathrm{unctr}}(d).
\end{equation}
\end{definition}

\begin{remark}
\bwang{
At time $t$, AeroSense takes as input all observed aircraft within the combined
scope
$
\mathbf{S}_t
=
\left\{
\mathbf{s}_{i,t}
\mid
\mathrm{Loc}(\mathbf{s}_{i,t}) \in \Omega(d)
\right\}.
$
Thus, $d$ controls the amount of surrounding traffic context used by the model, and complete surrounding airspace coverage is a configurable deployment setting for ATC rather than a prerequisite for complete surveillance. In our primary experiments, $d=+\infty$, such that all surrounding aircraft observable within $\Omega_{\mathrm{surv}}$ are included. Finite scopes $d=50$, $100$, and $200\,\mathrm{km}$ are further evaluated in the
}
\textit{\hyperref[ref:sensitivity]{Sensitivity Analysis}}.
\end{remark}

\begin{definition}[Prediction Task Modeling]
    Our objective is to predict the traffic volumes in the airspace of AP and AR, respectively, over a future prediction horizon $\Delta t$.  The model output is therefore denoted as

\begin{equation}
    \hat{\mathbf{Y}}_{(t,\Delta t)} = [\hat{y}_{AP}, \hat{y}_{AR}]^\top \in \mathbb{R}^2,
\end{equation}
\bwang{where $\hat{y}_{AP}$ and $\hat{y}_{AR}$ denote the predicted traffic volumes within the future interval $(t, t+\Delta t]$}, in the AP and AR airspaces, respectively. This study specially focuses on the 15-minute-ahead prediction (i.e., $\Delta t = 15\,\mathrm{min}$), which represents a practically tactical short-term prediction horizon in ATM. The proposed framework is readily applicable to longer prediction horizons (e.g., $\Delta t = 30\,\mathrm{min}$ or $90\,\mathrm{min}$), as discussed in Section~\ref{ref:dis}. 

The \textit{Prediction Task Modeling} is formulated as learning a mapping

\begin{equation}
f_\theta:\mathbf{S}_t \rightarrow \mathbb{R}^{2},
\end{equation}

such that

\begin{equation}
\hat{\mathbf{Y}}_{(t,\Delta t)}
\overset{\mathrm{def}}{=}
f_\theta(\mathbf{S}_t).
\end{equation}

When the prediction horizon $\Delta t$ is fixed, for notational simplicity, we abbreviate the notation as

\begin{equation}
\hat{\mathbf{Y}}_{t}
\overset{\mathrm{def}}{=}
f_\theta(\mathbf{S}_t).
\end{equation}
\end{definition}

\begin{remark}
    This formulation differs from conventional time-series-based flow forecasting methods, which operate on historical flow sequences. Instead, it directly maps the instantaneous airspace situation to future traffic flow. The central challenge is to construct a model $f_\theta$ that can effectively \textit{learn representations from the complex states of all aircraft in the set $\mathbf{S}_t$, while simultaneously operating on its variable cardinality.} Since the number of aircraft changes dynamically over time (i.e., $N_t \neq N_{t+1}$ in general; see Remark~\ref{ref:rm1}), the model must naturally accommodate varying set sizes of $\mathbf{S}_t$ without relying on fixed-input model structures. This requirement challenges conventional architectures such as CNNs, MLPs, and standard Transformers, which typically assume fixed-size inputs.
\end{remark}

\begin{definition}[Ground Truth]
The ground truth traffic flow $\mathbf{Y}_{(t,\Delta t)}$ is derived from aircraft trajectories obtained during the target horizon $(t, t+\Delta t]$. For a given query time $t$ and prediction horizon $\Delta t$, the  \textit{Ground Truth} for the airspace AR and AP is formally defined as

\begin{equation}\label{eq:yar}
    y_{AR}^t = y_{AR}^{(t,\Delta t]} \overset{\text{def}}{=}
    \left|
    \left\{
    i \;\middle|\; \mathrm{Loc}_\tau(i) \in AR,  \exists \tau \in (t, t+\Delta t]
    \right\}
    \right|,
\end{equation}

\begin{equation}\label{eq:yap}
    y_{AP}^t=y_{AP}^{(t,\Delta t]} \overset{\text{def}}{=}
    \left|
    \left\{
    i \;\middle|\; \mathrm{Loc}_\tau(i) \in AP, \exists \tau \in (t, t+\Delta t]
    \right\}
    \right|.
\end{equation}
Here, $\mathrm{Loc}_\tau(i)$ denotes the spatial location of aircraft $i$ at time $\tau$. In other words, each aircraft is counted once if it appears in the target airspace  at any time within $(t, t+\Delta t]$. The operator $|\cdot|$ denotes set cardinality.
\end{definition}

\begin{remark}
\bwang{The target in Eqs.~\ref{eq:yar}--\ref{eq:yap} measures \emph{interval regional demand}: each aircraft contributes once if it appears in the target airspace during the
forecast interval. It therefore characterizes traffic demand rather than
aircraft dwell time or simultaneous occupancy.}
\end{remark}

\subsection{Dataset and evaluation metrics}
\subsubsection{Dataset and missing value preprocessing}\label{sec:dataset}

\textbf{Dataset.} A large-scale real-world trajectory dataset was collected from the TA of an international airport, covering the period from March 1 to October 31, 2025. The dataset is derived from operational surveillance records provided by the local  ATC authority, where Secondary Surveillance Radar (SSR) \citep{gui2020machine} and ADS-B measurements are fused into unified aircraft trajectories. The fused trajectory records are broadcast to ATC at a temporal resolution of 4 seconds. 

Each trajectory record contains multi-source aircraft information, including temporal attributes, aircraft identifiers, departure and destination information, spatial positions, kinematic states, control-related variables, and sector-level operational labels. Collectively, these attributes characterize the instantaneous aircraft status from multiple perspectives, including where the aircraft is located, how it is moving, what control-intent cues it may follow, and which operational region it belongs to. 

Based on these raw attributes, we further construct aircraft-level state representations that encode location, motion dynamics, control-intent cues, boundary interactions, and temporal context for the subsequent state-to-flow prediction task (detailed in Section~\ref{sec:physics_features}).

\aqthenbw{\textbf{Missing value preprocessing.}
Due to occasional surveillance signal loss, aircraft observations may be missing at individual timestamps. At the 4-s sampling resolution of ADS-B, the overall missing rate is only 0.0381\%.  Among missing values, 90.83\% involve a single missing timestamp  (within 4 seconds) , 8.42\% involve two consecutive timestamps (within 8 seconds), and only 0.75\% involve three consecutive timestamps (within 12 seconds); no longer consecutive missing timestamps are observed. 
These statistics indicate that missing ADS-B observations are rare and predominantly short-lived. We therefore apply simple forward filling: when an observation is missing at time $t$, the aircraft's most recent valid state is carried forward.
}
{This preprocessing pipeline yields complete airspace situation records without missing values. In total, the resulting dataset comprises 224,904 samples, denoted by $\{(\mathbf{S}_t^{(k)}, \mathbf{Y}_t^{(k)})\}^{N}_{k=1}$ with $N = 224{,}904$. Following established practice in time-series predictor evaluation, we preserve temporal order and reserve the most recent observations for out-of-sample evaluation \citep{bergmeir2012crossvalidation}. Specifically, the samples are split chronologically into training, validation, and test sets in an 8:1:1 ratio, resulting in 179,923 training samples, 22,490 validation samples, and 22,490 test samples, respectively.\label{ref:testset}}

\subsubsection{Evaluation metrics}
We develop the model on the training and validation sets and evaluate its performance on the test set using four complementary metrics: mean absolute error (MAE), root mean squared error (RMSE), weighted absolute percentage error (WAPE), and coefficient of determination ($R^2$).Following best-practice guidance that forecasting performance should be assessed using valid and suitable metrics selected according to the characteristics of the data and the evaluation objective, we report multiple metrics to provide complementary perspectives on model performance \citep{hewamalage2023forecast}. \binwang{MAE measures the average magnitude of prediction errors and provides an intuitive assessment of forecasting accuracy. RMSE assigns a larger penalty to large deviations and is therefore more sensitive to extreme prediction errors. WAPE evaluates the relative forecasting error normalized by the total traffic volume, enabling fair comparisons across airspaces with different traffic scales.} Finally, $R^2$ quantifies the proportion of variance explained by the model and measures the overall goodness-of-fit. Let $\mathcal{D}_{test}=\{(\mathbf{S}_{t}^{(k)},\mathbf{Y}_{t}^{(k)})\}_{k=1}^{N}$ denote the test dataset with $N$ samples, where $\mathbf{S}_{t}^{(k)}$ represents the input aircraft states for the $k$-th sample, and $\mathbf{Y}_{t}^{(k)}=[y_{AP}^k,y_{AR}^k]^\top$ is the ground truth. Let $\hat{\mathbf{Y}}_{t}^{(k)}=[\hat{y}_{AP}^k,\hat{y}_{AR}^k]^\top$ denote the corresponding model prediction. Since the prediction results are reported separately for the AP and AR airspaces, the evaluation metrics are computed independently for each airspace $\mathcal{A}$ as:

\begin{equation}
\mathrm{MAE}_{\mathcal{A}}=\frac{1}{N}\sum_{k=1}^{N}\left|\hat{y}_\mathcal{A}^k-y_\mathcal{A}^k\right|
\end{equation}
\begin{equation}
\mathrm{RMSE}_{\mathcal{A}}=\sqrt{\frac{1}{N}\sum_{k=1}^{N}\left(\hat{y}_\mathcal{A}^k-y_\mathcal{A}^k\right)^2}
\end{equation}
\begin{equation}{
\mathrm{WAPE}_{\mathcal{A}}=
\frac{\sum_{k=1}^{N}\left|\hat{y}_{\mathcal{A}}^{k}-y_{\mathcal{A}}^{k}\right|}
{\sum_{k=1}^{N}\left|y_{\mathcal{A}}^{k}\right|}\times 100\%}
\end{equation}
\begin{equation}
R_{\mathcal{A}}^{2}=1-\frac{\sum_{k=1}^{N}\left(\hat{y}_\mathcal{A}^k-y_\mathcal{A}^k\right)^2}
{\sum_{k=1}^{N}\left(y_\mathcal{A}^k-\bar{y}_{\mathcal{A}}\right)^2}
\end{equation}
where $\bar{y}_{\mathcal{A}}=\frac{1}{N}\sum_{k=1}^{N}y_{\mathcal{A}}^{k}$ denotes the mean ground truth traffic volume of airspace $\mathcal{A}$ over the test set.

\subsection{Comparison baselines}
The {five} baseline groups are designed to isolate different sources of performance gain. {1) \aqthenbw{The naive baselines provide heuristic methods for reference.}} 2) Conventional time-series baselines evaluate whether direct microscopic state modeling outperforms classical macroscopic forecasting paradigms. 3) Augmented time-series baselines further test whether the gains can be explained solely by more physically grounded inputs. 4) \aqthenbw{ATM-specific forecasting baselines compare AeroSense with recent forecasting methods specifically developed for air traffic flow prediction.} 5) Finally, set-based baselines isolate the effect of permutation-invariant architectures from the proposed aircraft-level state modeling paradigm. 

\subsubsection{Naive baseline}
\binwang{\textbf{Lazy forecasting.} We include a simple persistence baseline, commonly referred to as \emph{lazy forecasting}, to provide a reference for the intrinsic difficulty of the prediction task. This method does not involve any model training and assumes that future traffic flow remains unchanged from the most recent observation. Specifically, given the current time $t$, the prediction for the future horizon is defined as $
\hat{\mathbf{Y}}_{t}
\overset{\mathrm{def}}{=}
\mathbf{Y}_{t-1}.
$}
\binwang{That is, the ground truth traffic flow $\mathbf{Y}_{t-1}$ observed during the previous forecasting interval is directly used as the prediction for the next interval. Despite its simplicity, this baseline is often surprisingly competitive in short-term forecasting problems due to the temporal continuity of traffic dynamics. Therefore, it serves as a useful reference point for evaluating whether more sophisticated models can effectively exploit predictive signals beyond simple persistence.}

\aqthenbw{\textbf{Additive forecasting.}
We also construct an additive baseline for comparison. This model independently predicts each aircraft’s future contribution to AP and AR and then aggregates these predictions into regional flow counts, providing an intuitive aircraft-level formulation of the counting task. The additive baseline adopts the same architecture and the same number of trainable parameters as AeroSense. The main difference is that it applies the classification head at the aircraft level and sums the resulting binary decisions to obtain the final flow count. Specifically, given the instantaneous aircraft set $\mathbf{S}_t$, the baseline uses the same architecture as AeroSense to obtain a contextualized representation $\mathbf{e}'_i$ for each aircraft (see Formula~\ref{ref:eprime}). For each target airspace $\mathcal{A}\in\{\mathrm{AP},\mathrm{AR}\}$, a region-specific classification head predicts
$
p_{i,\mathcal{A}}^{t}
=
\sigma\!\left(
g_{\mathcal{A}}^{\mathrm{add}}\!\left(\mathbf{e}'_i\right)
\right),
$
where $p_{i,\mathcal{A}}^{t}$ denotes the probability that aircraft $i$ will appear in airspace $\mathcal{A}$ during the prediction horizon $(t,t+\Delta t]$. Using a fixed threshold $\tau=0.5$, the predicted regional traffic flow is obtained by counting aircraft classified as present:
$
\hat{y}_{\mathcal{A}}^{t}
=
\sum_{i=1}^{N_t}
\mathbb{I}\!\left(
p_{i,\mathcal{A}}^{t}>\tau
\right).
$
}

\subsubsection{Conventional time-series  baselines} 
We compare AeroSense against several representative SOTA time-series forecasting models, all of which operate on aggregated historical flow sequences and represent the mainstream macroscopic forecasting paradigm. \bwang{For each baseline, we adopted the recommended hyperparameter configuration from the corresponding paper.}

\begin{itemize}
    \item \textbf{Autoformer \citep{wu2021autoformer}:} A decomposition-based Transformer that replaces self-attention with an auto-correlation mechanism to capture long-range series-wise dependencies.
    
    \item \textbf{FEDformer \citep{zhou2022fedformer}:} A frequency-enhanced Transformer that models global temporal patterns through seasonal-trend decomposition in the frequency domain.
    
    \item \textbf{TimesNet \citep{wu2023timesnet}:} A general-purpose time-series foundation model that transforms one-dimensional sequences into two-dimensional tensors to capture multi-scale temporal variations.
    
    \item \textbf{iTransformer \citep{liu2024itransformer}:} An inverted Transformer architecture that embeds the entire temporal sequence of each variate to model multivariate correlations.
    
    \item \textbf{DLinear \citep{zeng2023dlinear}:} A lightweight yet competitive linear forecasting model that decomposes time series into trend and residual components, serving as a strong linear baseline.

    \aqthenbw{\item \textbf{PhaseFormer \citep{niu2026phaseformer}:} An efficient time series forecasting model that represents periodic patterns through compact phase embeddings and captures cross phase dependencies using a lightweight routing mechanism.}

    \aqthenbw{\item \textbf{GTR \citep{ICLR2026_7a3284ce}:} A retrieval-based forecasting method that dynamically retrieves and aligns long-range periodic patterns from an adaptive global temporal embedding and integrates them with local historical observations.}

\end{itemize}

\subsubsection{Augmented time-series baselines} {To examine whether the observed performance gains arise merely from incorporating richer physical information, we further augment the conventional time-series models with physical features derived from aircraft states as below.}

{For a given target time $t$, we first construct a missing-value-free airspace situation following the procedure described in Section~\ref{sec:dataset}. Since time-series models cannot directly process variable-cardinality sets, we aggregate the aircraft states into a fixed-dimensional feature vector. Specifically, the aggregated states include: (i) aircraft counts in the AP, AR, and out-of-control airspace; (ii) summary statistics of boundary proximity to AP and AR, including mean, maximum, and quantiles; (iii) summary statistics of approach factor, including mean values and the proportion of aircraft with positive approach factor; (iv) kinematic statistics such as mean, standard deviation, and maximum ground speed, as well as mean, standard deviation, and absolute-mean vertical speed; (v) inclusion-indicator counts for AP and AR; and (vi) cyclical temporal features.}

{These aggregated vectors are then stacked into multivariate flow sequences and used as inputs to the augmented time-series models. In this way, the augmented baselines are provided with richer physically grounded states while remaining within the conventional multivariate time-series forecasting framework.}

\aqthenbw{\subsubsection{ATM-specific forecasting baselines}}

\aqthenbw{We further compare AeroSense with two recent air traffic flow methods specifically developed for ATM.}

\begin{itemize}
    \item \textbf{MTEFP \citep{ye2026multi}:}  \aqthenbw{A recent time series-based framework for air traffic flow prediction  that introduces multi-resolution temporal encoding to capture periodic correlations at different temporal scales and employs frequency-aware decomposition to separate short-term fluctuations from long-term traffic evolution.}

    \item \textbf{DFP-FG \citep{LIU2026111029}:} \aqthenbw{A trajectory-based air traffic flow prediction framework that incorporates directional flow pattern information derived from aircraft trajectories.}
\end{itemize}

\subsubsection{Set-based baselines}

{Finally, we investigate whether the observed performance gains can be attributed solely to the use of permutation-invariant set modeling. To this end, we consider two representative set-based architectures that can directly operate on aircraft-level inputs, enabling a direct comparison that isolates the effect of generic set-based modeling from the proposed airspace-aware design:}
{
\begin{itemize}
    \item \textbf{DeepSets} \citep{zaheer2017deep}: The first permutation-invariant model that aggregates independently processed elements via symmetric pooling.
    \item \textbf{SetTransformer} \citep{lee2019set}: An attention-based model designed to capture complex, higher-order interactions among elements.
\end{itemize}}
{Both models take the exact same dynamic set of aircraft states as \textit{AeroSense} as input.}

\subsection{Experimental configuration}
\aqthenbw{The dataset was chronologically split into training, validation, and test sets using an 8:1:1 ratio. All preprocessing statistics for feature and target normalization were estimated from the training set only. Model selection was based on the validation set, while the test set was reserved exclusively for final evaluation.}

\aqthenbw{AeroSense represents each aircraft using the 18 state variables defined in Section~\ref{sec:physics_features}. The aircraft-level encoder contains two fully connected layers with hidden dimension 128, each followed by batch normalization, ReLU, and dropout rate of $0.1$. Aircraft interactions are modeled with a 4-head self-attention layer, followed by two decoupled prediction heads for AP and AR traffic flow forecasting.}

\aqthenbw{Time-series baselines use the preceding 24 hours of observations sampled every 15 minutes, yielding an input length of 96, to predict traffic flow over the next 15 minutes. Timestamp embeddings are also provided for the input and forecast horizon. Conventional baselines use only historical regional flow sequences, while augmented variants additionally incorporate physical features aggregated from aircraft states. Architecture-specific hyperparameters follow the original corresponding studies \citep{wu2021autoformer,zhou2022fedformer,wu2023timesnet,liu2024itransformer,zeng2023dlinear,niu2026phaseformer,ICLR2026_7a3284ce}.}

\aqthenbw{All methods in our experimental pipeline were trained with Adam using an initial learning rate of $3\times10^{-4}$, weight decay of $1\times10^{-5}$, and batch size 32. Huber loss with $\delta=1.0$ was used as the objective. Training lasted up to 100 epochs, with the learning rate halved after 5 epochs without validation improvement and early stopping after 30 epochs. Gradients were clipped to a maximum norm of 1.0, and the checkpoint with the lowest validation loss was used for testing. All experiments were repeated with five random seeds, and results are reported as mean $\pm$ standard deviation. All experiments were implemented in Python 3.11 with PyTorch 2.5.1 and conducted on an Ubuntu 22.04 server equipped with one NVIDIA GeForce RTX 4090 GPU, an AMD EPYC 7302 CPU, and 128 GB RAM.} 


\subsection{Experimental results and quantitative analysis}\label{sec:results} \label{sec:experiments}
To systematically investigate whether microscopic aircraft-state modeling can transform air traffic flow prediction, we organize our experiments around three research questions (RQs):

\begin{itemize}
\item \textbf{RQ1:} \binwang{Can the proposed aircraft-level state-to-flow modeling paradigm provide more accurate traffic flow prediction than conventional time-series forecasting paradigms?} 
\item \textbf{RQ2:} Which microscopic aircraft-state factors and \binwang{module design} contribute most to  traffic dynamics?

\item \textbf{RQ3:} \binwang{Can the proposed AeroSense maintain predictive robustness under diverse traffic conditions, such as peak periods, missing aircraft, and aircraft-state noise?}

\end{itemize}
\subsubsection{Validation of the state-to-flow modeling paradigm (RQ1)}

\textbf{Comparison with the naive baseline.} {As shown in Table~\ref{tab:main_results}, lazy forecasting yields substantially larger prediction errors than the stronger learning-based models, with MAEs of 2.449 and 4.228 in the AP and AR airspaces, respectively. Its $R^2$ values are 0.801 in AP and 0.923 in AR, indicating that short-term traffic flow exhibits considerable temporal continuity. Nevertheless, the persistent gap between lazy forecasting and the proposed method demonstrates that future traffic evolution cannot be fully explained by simple persistence assumptions. Effective prediction requires modeling the underlying airspace dynamics beyond the most recent traffic observations.}

\aqthenbw{The additive forecasting baseline achieves MAEs of 3.218 and 3.191 in AP and AR, respectively, which are substantially higher than those of AeroSense (1.308 and 1.445). Although this baseline directly exploits aircraft-level state information, independently estimating the contribution of each aircraft and summing these predictions is insufficient to achieve comparable performance. This result further suggests that jointly modeling the aircraft set enables AeroSense to exploit microscopic aircraft-state information more effectively for regional traffic flow prediction.}

\textbf{Superior performance over the conventional time-series baselines.} \aqthenbw{Table~\ref{tab:main_results} reports the quantitative comparison between AeroSense and the baseline methods. Overall, AeroSense achieves the lowest prediction errors in both airspaces, with particularly pronounced gains in the high-volume airspace AR. Among the conventional time-series baselines, GTR achieves the strongest overall performance. Compared with GTR, AeroSense reduces MAE by approximately 14.1\% (1.308 vs.\ 1.523) and RMSE by 11.1\% (1.800 vs.\ 2.024) in AP. The improvement is substantially larger in AR, where AeroSense reduces MAE by approximately 46.0\% (1.445 vs.\ 2.675) and RMSE by 45.4\% (1.942 vs.\ 3.559). AeroSense also achieves the best WAPE in both airspaces and the best $R^2$ in AR, while attaining a comparable $R^2$ to GTR in AP (0.9415 vs.\ 0.9417).}

These results demonstrate that microscopic aircraft-level state modeling provides substantially higher predictive accuracy than macroscopic SOTA time-series models, especially for capturing the complex flow dynamics of high-volume airspace AR.

\begin{sidewaystable}[p]

\caption{
\aqthenbw{Comprehensive performance comparison with naive, conventional time-series,
augmented time-series, ATM-specific forecasting, and set-based baselines.
Results are reported as mean $\pm$ standard deviation over five random seeds where applicable.
Superscripts indicate statistical significance relative to AeroSense based on
two-sided paired $t$-tests over five independent runs for comparisons:
$^{*}p<0.05$, $^{**}p<0.01$, and $^{***}p<0.001$.}
}
\label{tab:main_results}

\centering
\fontsize{6}{7.1}\selectfont
\setlength{\tabcolsep}{1pt}

\begin{tabular}{@{}lcccccccc@{}}
\toprule
\multirow{2}{*}{\textbf{Model}}
& \multicolumn{4}{c}{\textbf{Airspace AP}}
& \multicolumn{4}{c}{\textbf{Airspace AR}} \\
\cmidrule(lr){2-5} \cmidrule(lr){6-9}
& \textbf{MAE}
& \textbf{RMSE}
& \textbf{WAPE}
& \textbf{R$^2$}
& \textbf{MAE}
& \textbf{RMSE}
& \textbf{WAPE}
& \textbf{R$^2$} \\

\midrule
\rowcolor[HTML]{F2F2F2}
\multicolumn{9}{l}{\textit{Naive baselines}} \\

Lazy forecasting
& $2.449^{***}$
& $3.314^{***}$
& $20.44^{***}\%$
& $0.801^{***}$
& $4.228^{***}$
& $5.669^{***}$
& $12.55^{***}\%$
& $0.923^{***}$ \\

Additive forecasting
& $3.218 \pm 0.042^{***}$
& $4.176 \pm 0.046^{***}$
& $26.890 \pm 0.353^{***}\%$
& $0.6853 \pm 0.0069^{***}$
& $3.191 \pm 0.074^{***}$
& $4.124 \pm 0.080^{***}$
& $9.484 \pm 0.221^{***}\%$
& $0.9595 \pm 0.0016^{***}$ \\

\midrule
\rowcolor[HTML]{F2F2F2}
\multicolumn{9}{l}{\textit{Conventional time-series baselines}} \\

Autoformer \citep{wu2021autoformer}
& $2.066 \pm 0.043^{***}$
& $2.646 \pm 0.055^{***}$
& $14.836 \pm 0.308^{***}\%$
& $0.9003 \pm 0.0042^{***}$
& $4.280 \pm 0.152^{***}$
& $5.486 \pm 0.182^{***}$
& $11.114 \pm 0.394^{***}\%$
& $0.9451 \pm 0.0037^{***}$ \\

FEDformer \citep{zhou2022fedformer}
& $1.925 \pm 0.202^{**}$
& $2.463 \pm 0.226^{**}$
& $13.819 \pm 1.447^{*}\%$
& $0.9131 \pm 0.0163$
& $4.394 \pm 1.081^{**}$
& $5.533 \pm 1.292^{**}$
& $11.411 \pm 2.808^{**}\%$
& $0.9418 \pm 0.0268^{*}$ \\

iTransformer \citep{liu2024itransformer}
& $1.784 \pm 0.213^{*}$
& $2.303 \pm 0.273^{*}$
& $12.810 \pm 1.530\%$
& $0.9237 \pm 0.0183$
& $3.245 \pm 0.315^{***}$
& $4.119 \pm 0.361^{***}$
& $8.428 \pm 0.818^{**}\%$
& $0.9689 \pm 0.0056^{**}$ \\

DLinear \citep{zeng2023dlinear}
& $1.664 \pm 0.001^{***}$
& $2.186 \pm 0.001^{***}$
& $11.949 \pm 0.008^{**}\%$
& $0.9320 \pm 0.0001^{**}$
& $3.026 \pm 0.004^{***}$
& $3.990 \pm 0.006^{***}$
& $7.858 \pm 0.012^{***}\%$
& $0.9710 \pm 0.0001^{***}$ \\

TimesNet \citep{wu2023timesnet}
& $1.551 \pm 0.013^{***}$
& $2.046 \pm 0.014^{***}$
& $11.133 \pm 0.095\%$
& $0.9404 \pm 0.0008$
& $2.783 \pm 0.032^{***}$
& $3.634 \pm 0.035^{***}$
& $7.227 \pm 0.083^{***}\%$
& $0.9759 \pm 0.0005^{***}$ \\

PhaseFormer \citep{niu2026phaseformer}
& $1.994 \pm 0.226^{**}$
& $2.635 \pm 0.323^{*}$
& $14.312 \pm 1.624^{*}\%$
& $0.9000 \pm 0.0248$
& $4.099 \pm 0.420^{***}$
& $5.457 \pm 0.614^{***}$
& $10.643 \pm 1.092^{**}\%$
& $0.9452 \pm 0.0113^{**}$ \\

GTR \citep{ICLR2026_7a3284ce}
& $1.523 \pm 0.005^{***}$
& \underline{$2.024 \pm 0.008^{***}$}
& \underline{$10.932 \pm 0.039\%$}
& $\mathbf{0.9417 \pm 0.0005}$
& $2.675 \pm 0.013^{***}$
& $3.559 \pm 0.008^{***}$
& $6.947 \pm 0.034^{***}\%$
& $0.9769 \pm 0.0001^{***}$ \\

\midrule
\rowcolor[HTML]{F2F2F2}
\multicolumn{9}{l}{\textit{Augmented time-series baselines}} \\

Augmented Autoformer
& $2.105 \pm 0.114^{***}$
& $2.814 \pm 0.131^{***}$
& $16.158 \pm 0.877^{**}\%$
& $0.8688 \pm 0.0124^{**}$
& $3.229 \pm 0.323^{***}$
& $4.204 \pm 0.393^{***}$
& $8.838 \pm 0.885^{**}\%$
& $0.9636 \pm 0.0070^{**}$ \\

Augmented FEDformer
& $1.895 \pm 0.044^{***}$
& $2.521 \pm 0.058^{***}$
& $14.551 \pm 0.334^{***}\%$
& $0.8949 \pm 0.0049^{***}$
& $3.010 \pm 0.092^{***}$
& $3.944 \pm 0.108^{***}$
& $8.238 \pm 0.253^{***}\%$
& $0.9682 \pm 0.0018^{***}$ \\

Augmented iTransformer
& $2.002 \pm 0.047^{***}$
& $2.632 \pm 0.060^{***}$
& $15.371 \pm 0.364^{***}\%$
& $0.8854 \pm 0.0052^{***}$
& $3.248 \pm 0.082^{***}$
& $4.205 \pm 0.099^{***}$
& $8.891 \pm 0.224^{***}\%$
& $0.9638 \pm 0.0017^{***}$ \\

Augmented DLinear
& $2.281 \pm 0.051^{***}$
& $2.888 \pm 0.071^{***}$
& $17.509 \pm 0.392^{***}\%$
& $0.8621 \pm 0.0067^{***}$
& $3.934 \pm 0.146^{***}$
& $4.951 \pm 0.194^{***}$
& $10.768 \pm 0.399^{***}\%$
& $0.9498 \pm 0.0038^{***}$ \\

Augmented TimesNet
& $1.992 \pm 0.100^{***}$
& $2.679 \pm 0.131^{**}$
& $15.292 \pm 0.764^{**}\%$
& $0.8811 \pm 0.0119^{**}$
& $3.094 \pm 0.134^{***}$
& $4.086 \pm 0.194^{***}$
& $8.468 \pm 0.366^{***}\%$
& $0.9658 \pm 0.0032^{***}$ \\

Augmented PhaseFormer
& $2.153 \pm 0.105^{***}$
& $2.772 \pm 0.072^{***}$
& $16.526 \pm 0.808^{***}\%$
& $0.8729 \pm 0.0066^{***}$
& $3.387 \pm 0.288^{***}$
& $4.289 \pm 0.315^{***}$
& $9.272 \pm 0.789^{***}\%$
& $0.9622 \pm 0.0055^{**}$ \\

Augmented GTR
& $1.818 \pm 0.021^{***}$
& $2.416 \pm 0.024^{***}$
& $13.960 \pm 0.162^{***}\%$
& $0.9035 \pm 0.0019^{***}$
& $2.943 \pm 0.046^{***}$
& $3.818 \pm 0.051^{***}$
& $8.056 \pm 0.125^{***}\%$
& $0.9702 \pm 0.0008^{***}$ \\

\midrule
\rowcolor[HTML]{F2F2F2}
\multicolumn{9}{l}{\textit{ATM-specific forecasting baselines}} \\

MTEFP \citep{ye2026multi}
& $1.813 \pm 0.014^{***}$
& $2.444 \pm 0.017^{***}$
& $15.165 \pm 0.117^{***}\%$
& $0.8922 \pm 0.0015^{***}$
& $3.173 \pm 0.012^{***}$
& $4.242 \pm 0.019^{***}$
& $9.439 \pm 0.035^{***}\%$
& $0.9572 \pm 0.0004^{***}$ \\

DFP-FG \citep{LIU2026111029}
& $2.055 \pm 0.029^{***}$
& $2.778 \pm 0.039^{***}$
& $17.171 \pm 0.246^{***}\%$
& $0.8608 \pm 0.0040^{***}$
& $3.685 \pm 0.197^{***}$
& $4.985 \pm 0.269^{***}$
& $10.954 \pm 0.587^{***}\%$
& $0.9407 \pm 0.0064^{***}$ \\

\midrule
\rowcolor[HTML]{F2F2F2}
\multicolumn{9}{l}{\textit{Set-based baselines}} \\

DeepSets \citep{zaheer2017deep}
& \underline{$1.520 \pm 0.021^{***}$}
& $2.062 \pm 0.030^{***}$
& $12.697 \pm 0.174^{***}\%$
& $0.9233 \pm 0.0022^{***}$
& \underline{$1.693 \pm 0.026^{***}$}
& \underline{$2.278 \pm 0.038^{***}$}
& \underline{$5.032 \pm 0.076^{***}\%$}
& \underline{$0.9877 \pm 0.0004^{***}$} \\

SetTransformer \citep{lee2019set}
& $1.694 \pm 0.192^{*}$
& $2.320 \pm 0.261^{*}$
& $14.157 \pm 1.607^{*}\%$
& $0.9019 \pm 0.0213$
& $2.735 \pm 0.673^{*}$
& $3.738 \pm 0.947^{*}$
& $8.130 \pm 2.000^{*}\%$
& $0.9650 \pm 0.0154^{*}$ \\

\midrule
\textbf{AeroSense (Ours)}
& $\mathbf{1.308 \pm 0.018}$
& $\mathbf{1.800 \pm 0.029}$
& $\mathbf{10.930 \pm 0.148\%}$
& $\underline{0.9415 \pm 0.0019}$
& $\mathbf{1.445 \pm 0.021}$
& $\mathbf{1.942 \pm 0.029}$
& $\mathbf{4.293 \pm 0.063\%}$
& $\mathbf{0.9910 \pm 0.0003}$ \\

\bottomrule
\end{tabular}

\end{sidewaystable}

\textbf{Comparison with augmented time-series baselines.}
\aqthenbw{Table~\ref{tab:main_results} also shows that adding aggregated physical-state features to conventional time-series models can improve forecasting performance, particularly for some models in the high-volume AR airspace. For example, the MAE of Autoformer decreases from 4.280 to 3.229 after augmentation, while FEDformer and PhaseFormer improve from 4.394 and 4.099 to 3.010 and 3.387, respectively. These improvements indicate that aggregated physical-state features carry useful information for dense traffic flow prediction, although they provide only coarse summaries of aircraft-level states. Meanwhile, the gains are not consistent across all architectures. For example, in AR, the MAE of DLinear increases from 3.026 to 3.934 after augmentation, while that of GTR increases from 2.675 to 2.943. The results in the AP airspace show a similar dependence on model architecture. Since AP has a lower traffic volume, its future flow can be more sensitive to the states of a relatively small number of individual aircraft. Under this condition, compressing microscopic aircraft states into fixed-dimensional statistics may smooth out important aircraft-level differences. Consequently, the benefit of augmentation is generally limited and varies across models, although certain models such as FEDformer still obtain modest improvements.}

\aqthenbw{{These results show that aircraft-state information is useful, especially in dense traffic conditions, but the way it is incorporated matters.} Coarse aggregation can improve some conventional time-series models, but does not provide consistent gains across different model architectures or airspaces. AeroSense avoids this limitation by modeling aircraft states directly as a variable-cardinality set, allowing it to preserve aircraft-level information and achieve stable performance in both AP and AR airspaces.}

\aqthenbw{\textbf{Comparison with ATM-specific forecasting baselines.}
AeroSense also consistently outperforms the two recent ATM-specific forecasting methods, MTEFP and DFP-FG, in both airspaces. Among them, MTEFP achieves the stronger overall performance, with MAEs of 1.813 in AP and 3.173 in AR. Compared with MTEFP, AeroSense reduces MAE by approximately 27.9\% (1.308 vs.\ 1.813) and RMSE by 26.4\% (1.800 vs.\ 2.444) in AP. The improvement becomes more pronounced in the high-volume AR airspace, where AeroSense reduces MAE by approximately 54.5\% (1.445 vs.\ 3.173) and RMSE by 54.2\% (1.942 vs.\ 4.242). These results demonstrate that directly modeling fine-grained aircraft-level states provides complementary predictive information beyond the time series-based and trajectory-based representations employed by existing SOTA ATM-specific forecasting methods.}

\textbf{Comparison with set-based baselines.}
The set-based baselines in Table~\ref{tab:main_results} demonstrate that directly modeling airspace situations as aircraft sets already yields strong performance. DeepSets outperforms most conventional time-series baselines, achieving MAEs of 1.520 and 1.693 in the AP and AR airspaces, respectively. This result suggests that set-based modeling is naturally suited to the variable-cardinality nature of airspace situations. Building upon \aqthenbw{the set-based formulation}, SetTransformer further introduces self-attention to capture interactions among aircraft, achieving MAEs of 1.694 in AP and 2.735 in AR. 

\aqthenbw{These results further motivate investigating whether explicitly modeling inter-aircraft interactions can provide additional benefits for traffic flow prediction.} Overall, AeroSense achieves the best performance, with MAEs of 1.308 in AP and 1.445 in AR. While the improvement over DeepSets is modest, it is consistent across both airspaces; the gain over SetTransformer is substantially larger, particularly in AR. This suggests that the effectiveness of AeroSense stems not only from adopting a set-based representation, but also from its aircraft-level state and \aqthenbw{the model architecture design.}

\begin{figure}[!hbt]
  \centering
  \includegraphics[width=\linewidth]{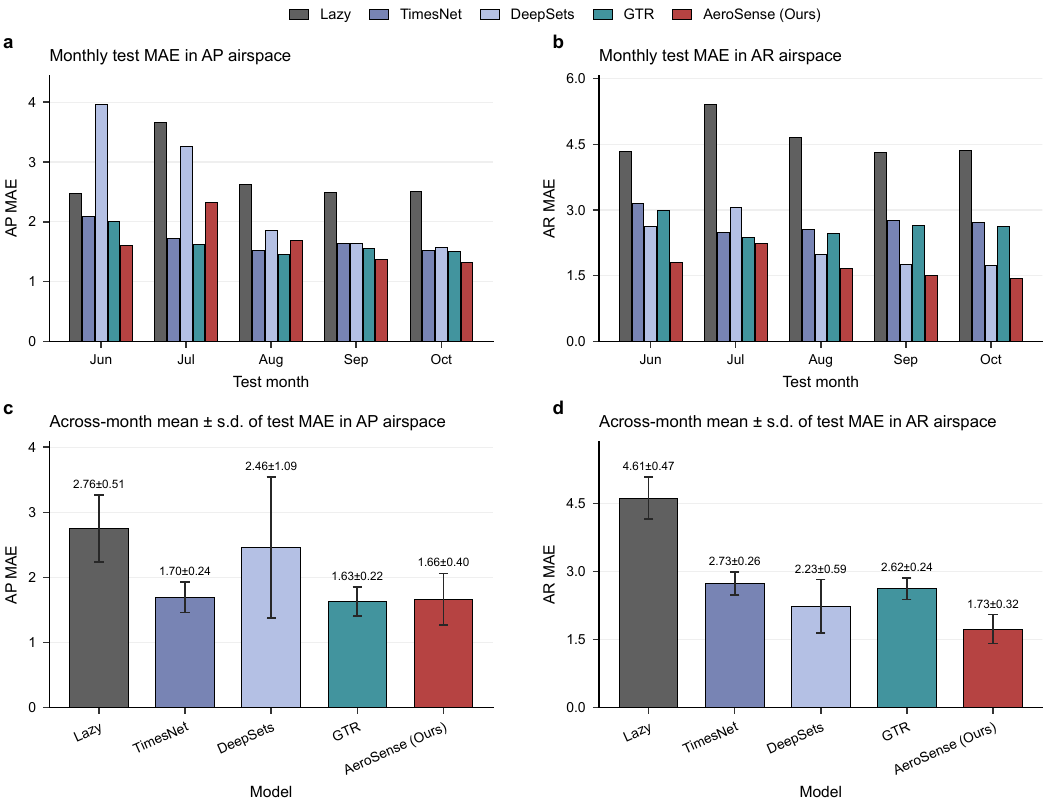}
  \caption{\textbf{Month-wise generalization performance.}
(a) AP and (b) AR prediction MAE when each month from June to October is independently used as the test set. 
(c) AP and (d) AR aggregated results reported as mean $\pm$ standard deviation of MAE across all monthly test sets.
\aqthenbw{AeroSense achieves the lowest across-month average MAE in AR and remains highly competitive in AP, demonstrating strong generalization under month-to-month distribution shifts.}}
  \label{fig:monthly_generalization}
\end{figure}

\textbf{Monthly test.} To evaluate robustness under temporal distribution shifts, we further perform a month-wise generalization study using the five-month subset from June to October 2025, where each month is used as the test set in turn and all models are re-trained on the remaining four months following the same experimental protocol. \aqthenbw{Despite month-to-month variations, AeroSense maintains consistently competitive performance across the different monthly test sets, with a particularly clear advantage in the AR airspace.} As shown in Fig.~\ref{fig:monthly_generalization}(a) and (b), prediction errors vary across months for all methods, reflecting the heterogeneous traffic patterns over time. Nevertheless, AeroSense achieves the lowest MAE in most months for both the AP and AR airspaces, \aqthenbw{with the only exceptions being July and August in the AP airspace, where GTR performs slightly better.} The advantage is particularly pronounced in the AR airspace, which exhibits higher traffic density and stronger temporal variability. 


\aqthenbw{Unlike Table \ref{tab:main_results}, where variability is measured on a fixed test set, Figs. ~\ref{fig:monthly_generalization} (c) and (d) report the mean and standard deviation across five monthly test sets, with each month evaluated separately as the held-out test period. They therefore characterize cross-month performance variability under changing traffic patterns. While the set-based baselines generally outperform conventional time-series and lazy forecasting methods, AeroSense and GTR achieve the most competitive performance. In the AP airspace, GTR attains a marginally lower average MAE than AeroSense (1.63 vs. 1.66). In contrast, in the AR airspace, AeroSense achieves the lowest average MAE of 1.73, substantially outperforming GTR (2.62) and the other baselines. Overall, these results indicate that the proposed state-to-flow paradigm maintains strong predictive accuracy and robustness across varying monthly traffic conditions.}

\subsubsection{Ablation study of AeroSense design (RQ2)}
To understand the effect of the key design choices in AeroSense, we conduct a comprehensive ablation study. The results are summarized in Table~\ref{tab:ablation}.

\clearpage

\begin{sidewaystable}[p]
  \caption{
  \aqthenbw{Ablation study results of AeroSense. Results are reported as mean $\pm$ standard deviation over five random seeds. \textbf{w/o} denotes the removal of a specific component. Superscripts indicate statistical significance relative to AeroSense based on two-sided paired $t$-tests over five independent runs for  comparisons: $^{*}p<0.05$, $^{**}p<0.01$, and $^{***}p<0.001$.}
  }
  \label{tab:ablation}
  \centering
  \fontsize{6}{7.1}\selectfont
  \setlength{\tabcolsep}{1pt}

  \begin{tabular}{@{}lcccccccc@{}}
    \toprule
    \multirow{2}{*}{\textbf{Variant}}
    & \multicolumn{4}{c}{\textbf{Airspace AP}}
    & \multicolumn{4}{c}{\textbf{Airspace AR}} \\
    \cmidrule(lr){2-5}
    \cmidrule(lr){6-9}
    & \textbf{MAE}
    & \textbf{RMSE}
    & \textbf{WAPE}
    & \textbf{R$^2$}
    & \textbf{MAE}
    & \textbf{RMSE}
    & \textbf{WAPE}
    & \textbf{R$^2$} \\
    \midrule

    \multicolumn{9}{l}{\textit{Effect of module design}} \\

    w/o Decoupled prediction heads
    & $1.323 \pm 0.017$
    & $1.805 \pm 0.024$
    & $11.06 \pm 0.14\%$
    & $0.9412 \pm 0.0016$
    & $1.492 \pm 0.034^{*}$
    & $1.994 \pm 0.039^{*}$
    & $4.43 \pm 0.10\%^{*}$
    & $0.9905 \pm 0.0004$ \\

    w/o Masked self-attention
    & $1.485 \pm 0.056^{*}$
    & $1.994 \pm 0.065^{*}$
    & $12.41 \pm 0.47\%^{*}$
    & $0.9282 \pm 0.0047^{*}$
    & $1.848 \pm 0.095^{**}$
    & $2.493 \pm 0.144^{*}$
    & $5.49 \pm 0.28\%^{**}$
    & $0.9852 \pm 0.0017^{*}$ \\

    \midrule
    \multicolumn{9}{l}{\textit{Effect of aircraft-state representation}} \\

    w/o State of boundary interactions
    & $1.462 \pm 0.024^{**}$
    & $1.996 \pm 0.035^{**}$
    & $12.22 \pm 0.20\%^{**}$
    & $0.9281 \pm 0.0026^{*}$
    & $1.890 \pm 0.240^{*}$
    & $2.530 \pm 0.318^{*}$
    & $5.62 \pm 0.71\%^{*}$
    & $0.9846 \pm 0.0041$ \\

    w/o State of aircraft location
    & $1.353 \pm 0.024^{*}$
    & $1.845 \pm 0.035$
    & $11.30 \pm 0.20\%^{*}$
    & $0.9385 \pm 0.0023$
    & $1.630 \pm 0.026^{***}$
    & $2.187 \pm 0.040^{**}$
    & $4.84 \pm 0.08\%^{***}$
    & $0.9886 \pm 0.0004^{**}$ \\

    w/o State of aircraft kinematic
    & $1.326 \pm 0.021^{*}$
    & $1.832 \pm 0.029^{*}$
    & $11.08 \pm 0.18\%^{*}$
    & $0.9394 \pm 0.0019^{*}$
    & $1.682 \pm 0.083^{*}$
    & $2.265 \pm 0.115^{*}$
    & $5.00 \pm 0.25\%^{*}$
    & $0.9878 \pm 0.0012^{*}$ \\

    w/o State of temporal context
    & $1.404 \pm 0.016^{**}$
    & $1.923 \pm 0.023^{**}$
    & $11.73 \pm 0.13\%^{**}$
    & $0.9333 \pm 0.0016^{**}$
    & $1.508 \pm 0.007^{*}$
    & $2.012 \pm 0.009^{*}$
    & $4.48 \pm 0.02\%^{*}$
    & $0.9904 \pm 0.0001^{*}$ \\

    w/o State of controlling intent
    & $1.388 \pm 0.043^{*}$
    & $1.904 \pm 0.043^{*}$
    & $11.59 \pm 0.36\%^{*}$
    & $0.9345 \pm 0.0030^{*}$
    & $1.606 \pm 0.090^{*}$
    & $2.155 \pm 0.113^{*}$
    & $4.77 \pm 0.27\%^{*}$
    & $0.9889 \pm 0.0012$ \\

    Minimal Physical State
    & $1.487 \pm 0.040^{**}$
    & $2.028 \pm 0.061^{**}$
    & $12.42 \pm 0.33\%^{**}$
    & $0.9257 \pm 0.0045^{*}$
    & $1.770 \pm 0.048^{***}$
    & $2.356 \pm 0.061^{***}$
    & $5.26 \pm 0.14\%^{***}$
    & $0.9868 \pm 0.0007^{***}$ \\

    \midrule
    \multicolumn{9}{l}{\textit{Effect of pooling strategy}} \\

    SumPooling$\rightarrow$MeanPooling
    & $1.714 \pm 0.100^{**}$
    & $2.333 \pm 0.133^{**}$
    & $14.32 \pm 0.83\%^{**}$
    & $0.9016 \pm 0.0111^{*}$
    & $3.896 \pm 0.152^{***}$
    & $5.328 \pm 0.221^{***}$
    & $11.58 \pm 0.45\%^{***}$
    & $0.9324 \pm 0.0056^{***}$ \\

    SumPooling$\rightarrow$MaxPooling
    & $1.899 \pm 0.120^{**}$
    & $2.578 \pm 0.151^{**}$
    & $15.87 \pm 1.00\%^{**}$
    & $0.8797 \pm 0.0140^{**}$
    & $5.410 \pm 0.307^{***}$
    & $6.838 \pm 0.349^{***}$
    & $16.08 \pm 0.91\%^{***}$
    & $0.8886 \pm 0.0114^{***}$ \\

    \midrule
    \textbf{AeroSense (Our full model)}
    & $\mathbf{1.308 \pm 0.018}$
    & $\mathbf{1.800 \pm 0.029}$
    & $\mathbf{10.93 \pm 0.15\%}$
    & $\mathbf{0.9415 \pm 0.0019}$
    & $\mathbf{1.445 \pm 0.021}$
    & $\mathbf{1.942 \pm 0.029}$
    & $\mathbf{4.29 \pm 0.06\%}$
    & $\mathbf{0.9910 \pm 0.0003}$ \\

    \bottomrule
  \end{tabular}
\end{sidewaystable}

\clearpage

\textbf{Effect of module design.} As shown in Table~\ref{tab:ablation}, the \textit{w/o decoupled prediction heads} variant consistently underperforms the full model, increasing the MAE from 1.308 to 1.323 in the AP airspace and from 1.445 to 1.492 in the AR airspace. This result suggests that AP and AR flow prediction require different output mappings due to their distinct traffic densities and aircraft relevance patterns. A shared prediction head couples the two targets and may cause negative transfer when the informative aircraft-state cues differ across airspaces. By using decoupled heads, AeroSense enables airspace-specific prediction branches and reduces interference between the two tasks.

{Furthermore, the inclusion of the \textit{masked self-attention} mechanism proves crucial for capturing implicit inter-aircraft dependencies. As indicated in Table~\ref{tab:ablation}, removing this module (\textit{w/o masked self-attention}) leads to a noticeable decline in predictive performance, increasing the MAE from 1.308 to 1.485 in the AP airspace, and from 1.445 to 1.848 in the AR airspace. Without self-attention mechanism, the network processes each aircraft's state in isolation, thereby failing to capture the structural relationships and interactions among the valid aircraft. By leveraging masked self-attention, AeroSense dynamically evaluates the relative influence between aircraft pairs, identifying influential aircraft while suppressing irrelevant ones, which is essential for accurately modeling the airspace situation.}

\begin{figure}[!hbt]
  \centering
  \includegraphics[width=\linewidth]{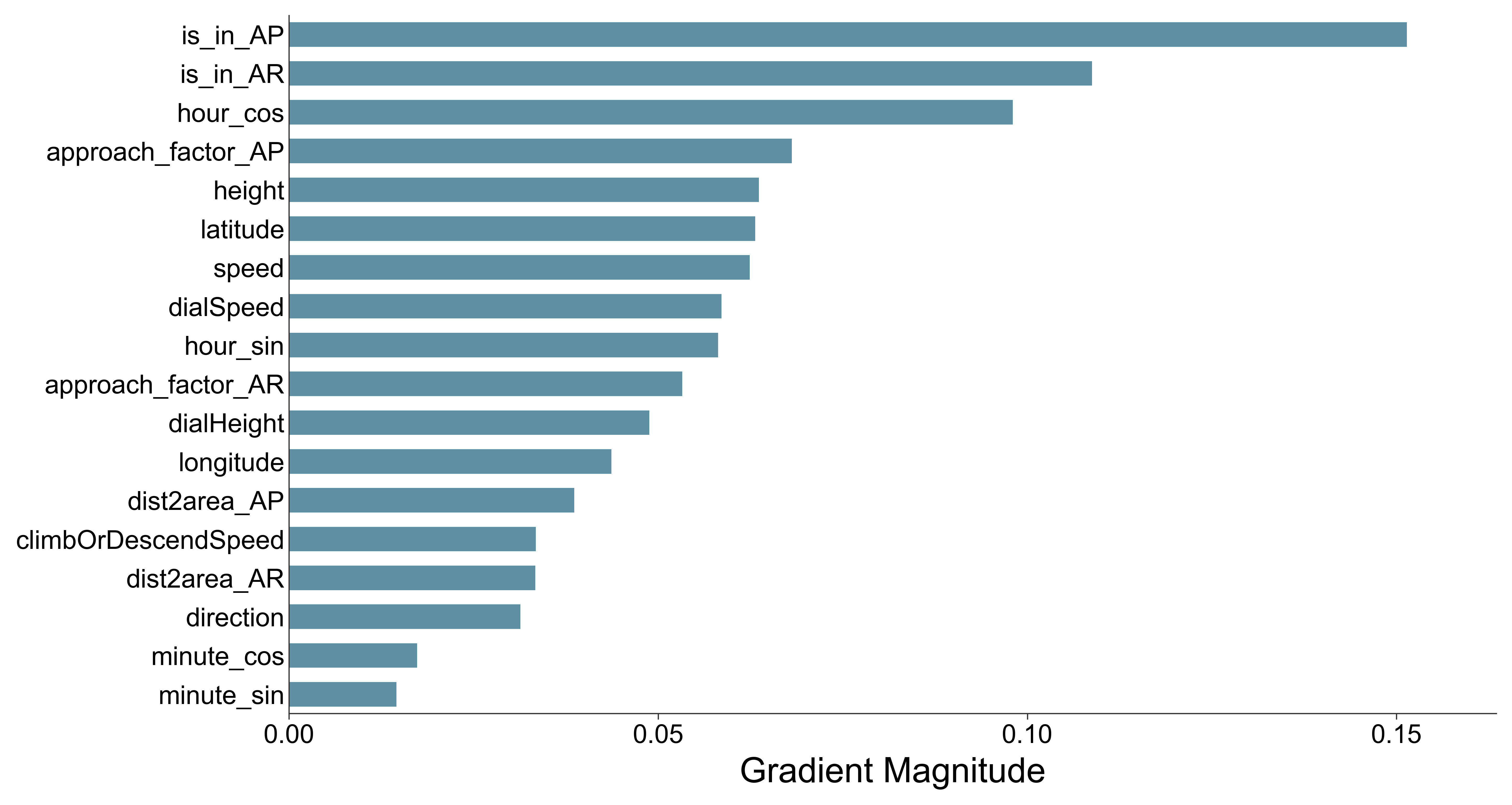}
  \caption{\textbf{Importance of the proposed aircraft states.} The y-axis lists the proposed microscopic aircraft states. The x-axis represents the gradient magnitude, which measures the sensitivity of the model prediction to each state.}
  \label{fig:feature_imp}
\end{figure}

\textbf{Effect of aircraft-state representation.} As shown in Table~\ref{tab:ablation}, removing the boundary interactions states ($w/o~State~of~boundary~interactions$) leads to a clear performance drop, with the MAE increasing from 1.308 to 1.462 in the AP airspace and from 1.445 to 1.890 in the AR airspace. This decline confirms that merely GPS coordinates alone are insufficient to capture complex boundary dynamics. The conclusion is further supported by the state importance analysis in Fig.~\ref{fig:feature_imp}, where the \textit{approach factor} and \textit{airspace inclusion indicator} rank among the most influential states. Taken together, these findings suggest that AeroSense relies primarily on relationships between aircraft and airspace, rather than merely memorizing absolute GPS coordinates.

{To further examine the necessity of the designed physics-informed states, we evaluate a \textit{Minimal Physical State} variant that retains only basic location and kinematic states. 
As reported in Table~\ref{tab:ablation}, this simplified variant still achieves competitive performance (AP MAE 1.487, AR MAE 1.770), remaining substantially stronger than conventional time-series baselines in both the AR and AP airspaces. 
This finding suggests that the primary performance gain originates  from the proposed state-to-flow modeling paradigm itself, while the full situation-aware  state representation provides additional performance improvements.}

\textbf{Effect of pooling strategy.}
{The choice of pooling strategy is a fundamental structural design in AeroSense rather than a simple implementation detail. As shown in Table~\ref{tab:ablation}, replacing the default \textit{SumPooling} operator (defined in Eq.~\eqref{eq:pooling}) with alternative aggregation strategies leads to substantial performance degradation. In the high-density AR airspace, adopting \textit{MeanPooling} and \textit{MaxPooling} increases the MAE from 1.445 to 3.896 and 5.410, respectively. Similar trends are also observed in the AP airspace.} 

{\textit{Why is the pooling strategy so influential?} The key reason lies in the intrinsic nature of air traffic flow prediction. Traffic volume is fundamentally an accumulative quantity over aircraft: it depends on the number of relevant aircraft and their individual contributions to future flow, rather than on the averaged state of a typical aircraft or the response of a single dominant aircraft. Consequently, the global airspace representation should remain sensitive to the number of valid aircraft contained in the input set.}

{Consider two traffic situations with similar average aircraft states but substantially different aircraft counts. Under \textit{MeanPooling}, these situations may produce nearly identical global representations despite corresponding to very different future traffic volumes. In contrast, \textit{MaxPooling} introduces another limitation by preserving only the strongest activation along each feature dimension while discarding contributions from the remaining aircraft. Such behavior is also unsuitable for flow prediction, where multiple moderately relevant aircraft may jointly determine future traffic demand.}

{By comparison, \textit{SumPooling} accumulates the representations of all valid aircraft and therefore better aligns with the counting nature of the prediction task. Under this formulation, each aircraft can be interpreted as contributing a certain amount of evidence toward future AP or AR traffic flow. By simultaneously preserving aircraft-level state information and the scale of the aircraft set, \textit{SumPooling} provides a more appropriate global representation for air traffic flow prediction.}

\subsubsection{\binwang{Robustness under diverse traffic conditions (RQ3)}}

\binwang{\textbf{Robustness under temporally heterogeneous traffic periods.}} Air traffic flows exhibit pronounced temporal heterogeneity, with substantial demand surges during morning and evening peak periods. To systematically evaluate the robustness of AeroSense under such volatile conditions, we conduct a dayparting multi-object evaluation by analyzing the prediction errors across different time periods over a full 24-hour cycle. Specifically, the day is partitioned into twelve non-overlapping 2-hour intervals, each treated as an independent evaluation objective.

\begin{figure}[p]
  \centering
  \includegraphics[ width=\linewidth, height=0.78\textheight, keepaspectratio ]
  {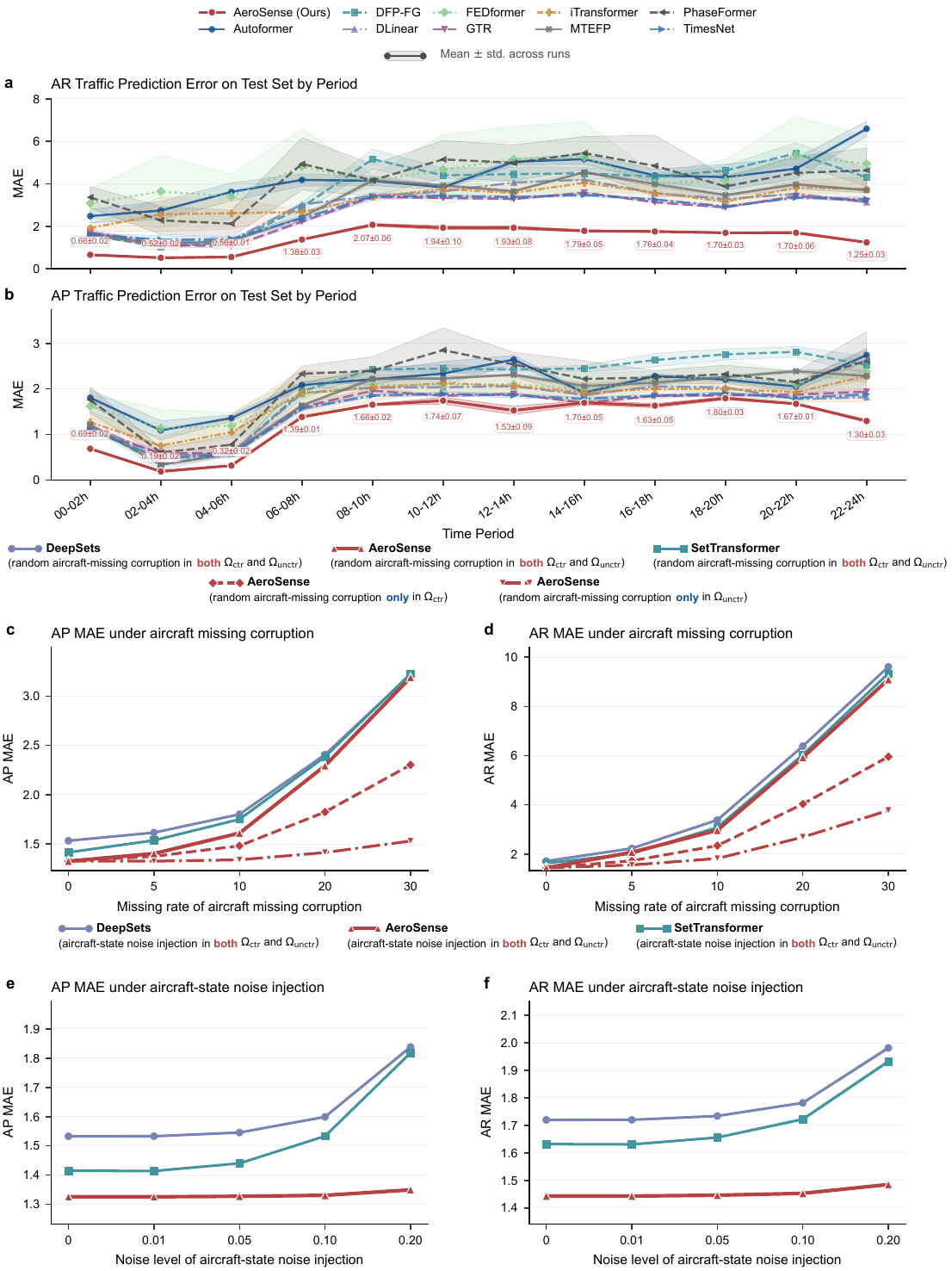}
  \caption{\aqthenbw{
\textbf{Robustness evaluation under temporal traffic heterogeneity, aircraft missing corruption, and aircraft-state noise injection.}
(a,b) Prediction MAE across twelve non-overlapping two-hour periods for the AR and AP airspaces, respectively. Each point represents the average prediction error within the corresponding time interval. AeroSense consistently achieves lower errors across most periods, particularly during high-traffic daytime operations characterized by stronger traffic fluctuations.
(c,d) Robustness to aircraft missing corruption in the AP and AR airspaces, respectively. The missing rate denotes the proportion of aircraft randomly removed from the input set at test time. In addition to random corruption in both $\Omega_{\mathrm{ctr}}$ and $\Omega_{\mathrm{unctr}}$, two region-specific settings are considered for AeroSense, where aircraft are removed only from $\Omega_{\mathrm{ctr}}$ or only from $\Omega_{\mathrm{unctr}}$. AeroSense exhibits substantially slower performance degradation than competing set-based baselines as the missing rate increases.
(e,f) Robustness to aircraft-state noise injection in the AP and AR airspaces, respectively. The noise level $\sigma$ denotes the standard deviation of Gaussian perturbations added to continuous aircraft-state variables at test time. AeroSense remains considerably more stable than DeepSets and SetTransformer across all noise levels, indicating strong resilience to measurement uncertainty.
}}
  \label{fig:robustness_combined}
\end{figure}

{As shown in Fig.~\ref{fig:robustness_combined}(a) and Fig.~\ref{fig:robustness_combined}(b), AeroSense consistently achieves low MAE across all time intervals in both the AP and AR airspaces. In particular, \textit{AeroSense attains Pareto-optimal performance in the AR airspace}, forming a dominant error frontier that is not surpassed by any baseline under this dayparting multi-objective evaluation. The advantage of AeroSense is especially pronounced during high-volatility periods, such as the morning congestion peak (08{:}00--10{:}00). Whereas macroscopic time-series baselines (e.g., Autoformer and FEDformer) exhibit substantial degradation with sharp error spikes. For instance, their MAE surges to approximately 4.0 and 3.4 in the AR airspace during this morning peak. AeroSense maintains a smoother and more stable error profile, keeping its MAE at a significantly lower level of roughly 2.0. A similar contrast is observed during the late-night period (22{:}00--24{:}00); while the Autoformer error spikes dramatically to nearly 6.6 in the AR and 2.8 in the AP, AeroSense remains highly robust, with its MAE dropping below 1.2 and 1.3, respectively. This robustness demonstrates that directly modeling instantaneous aircraft states allows the model to rapidly adapt to highly dynamic traffic flow transitions, providing more reliable predictions during sudden demand surges.}

\textbf{Robustness to aircraft missing corruption and aircraft-state noise injection.}
\aqthenbw{Since real-world ADS-B data may be affected by intermittent observation missingness and measurement noise in aircraft states, we evaluate the robustness of AeroSense under two types of simulated perturbations: missing observations and aircraft-state noise injection. These perturbations are intended as controlled stress tests rather than exhaustive simulations of specific operational surveillance failures.} All perturbations are applied only to the test set (22,490 samples; see Section~\ref{ref:testset}), while the training data and model parameters remain unchanged.

\begin{itemize}
    \item \binwang{\textit{Aircraft missing corruption}. We consider three missing scenarios: (1) aircraft are randomly removed from both the controlled region $\Omega_{\mathrm{ctr}}$ and the surrounding region $\Omega_{\mathrm{unctr}}$; (2) aircraft are removed only from $\Omega_{\mathrm{ctr}}$; and (3) aircraft are removed only from $\Omega_{\mathrm{unctr}}$. For each scenario, aircraft are randomly discarded with missing rates of $0\%$, $5\%$, $10\%$, $20\%$, and $30\%$.} \binwang{As shown in Fig.~\ref{fig:robustness_combined}(c) and (d), increasing the missing rate consistently degrades the performance of all models, which is expected because the corrupted input provides a less complete description of the airspace situation. Nevertheless, AeroSense remains the most robust method across all missing rates, consistently outperforming DeepSets and SetTransformer. Moreover, comparing the two region-specific settings reveals that missing aircraft in the controlled region $\Omega_{\mathrm{ctr}}$ causes a substantially larger increase in prediction error than missing aircraft in $\Omega_{\mathrm{unctr}}$. This observation suggests that aircraft currently under ATC supervision contribute more directly to future traffic evolution. At the same time, the performance degradation observed when aircraft are removed from $\Omega_{\mathrm{unctr}}$ indicates that surrounding out-of-control traffic still provides valuable contextual information for future flow prediction.}

    \item \binwang{\textit{Aircraft-state noise injection}. We further evaluate robustness to measurement noise by perturbing the state features of each aircraft in the test set. This setting simulates uncertainty in ADS-B measurements while preserving the original aircraft composition and traffic volume. Specifically, for the continuous component vector $\mathbf{s}$ of each aircraft state $\mathbf{s}_i^t$ (see Eq.~\ref{ref:istate}), the perturbed state $\tilde{\mathbf{s}}$ is generated as
$
\tilde{\mathbf{s}}=\mathbf{s}+\boldsymbol{\epsilon},
\,
\boldsymbol{\epsilon}\sim\mathcal{N}(\mathbf{0},\sigma^2\mathbf{I}),
$
where $\sigma$ controls the noise level in the normalized feature space. We consider $\sigma\in\{0,0.01,0.05,0.10,0.20\}$. Non-continuous state variables are left unchanged. In total, $8$ out of $18$ input state variables are subjected to noise perturbation. Fig.~\ref{fig:robustness_combined}(e) and Fig.~\ref{fig:robustness_combined}(f) report the results under different noise levels. As the noise intensity increases, the MAE of all methods gradually rises. Nevertheless, AeroSense exhibits the strongest robustness, with a substantially smaller increase in MAE than DeepSets and SetTransformer across all noise levels.}
\end{itemize}

These results suggest that AeroSense is less sensitive to perturbations in individual aircraft and can maintain reliable flow prediction under realistic sensing uncertainty.

\phantomsection
\label{ref:sensitivity}

\binwang{\textbf{Sensitivity to the scope of out-of-control airspace.}
We investigate the sensitivity of AeroSense to the scope of the out-of-control airspace, denoted by $\Omega_{\mathrm{unctr}}$ and parameterized by the hyperparameter $d$ in Eq.~\ref{ref:d}. Specifically, we vary $d \in \{0,50,100,200,+\infty\}$, where $d=0$ considers only aircraft within the controlled airspace $\Omega_{\mathrm{ctr}}$. The settings $d\in\{50,100,200\}$ progressively incorporate aircraft located within $d$ km outside the boundary of $\Omega_{\mathrm{ctr}}$, while $d=+\infty$ corresponds to the full surveillance coverage available to the ATM authority. Notably, the setting $d=100$ km approximately matches the maximum distance that an aircraft can travel within the 15-minute prediction horizon and therefore serves as a practically meaningful operating point.
}

\begin{figure}[!hbt]
  \centering
  \includegraphics[width=\linewidth]{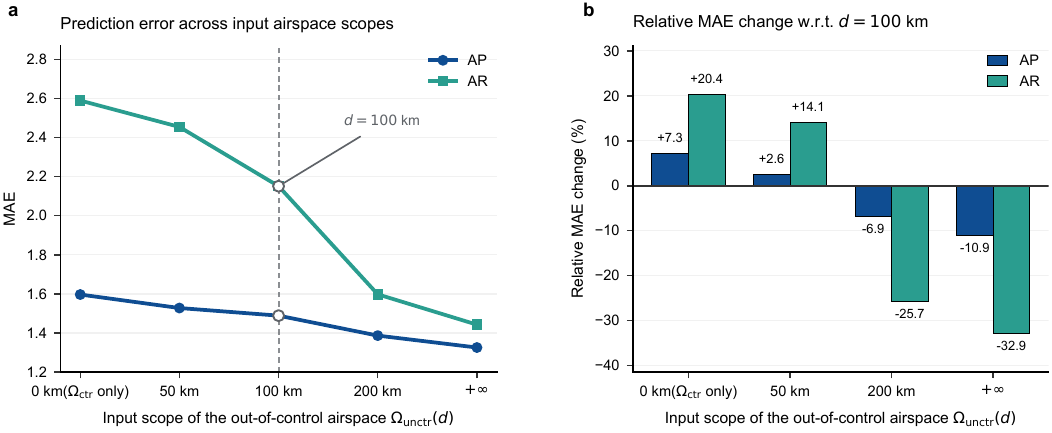}
  \caption{\binwang{\textbf{Sensitivity to the scope of the out-of-control airspace.}
(a) Prediction MAE of AeroSense under different spatial scopes of the out-of-control airspace $\Omega_{\mathrm{unctr}}$. Including a broader surrounding airspace generally decreases prediction MAE. 
(b) Relative MAE change with respect to the reference setting $d=100$ km. The full-surveillance setting ($d=+\infty$) achieves the best gains, highlighting the importance of incorporating broader traffic context beyond the controlled airspace.}
}
  \label{fig:spatial_scope_sensitivity}
\end{figure}

\binwang{As shown in Fig.~\ref{fig:spatial_scope_sensitivity}, enlarging the scope of $\Omega_{\mathrm{unctr}}$ consistently improves prediction performance in both AP and AR. Compared with the reference setting of $d=100$ km, the full-surveillance setting ($d=+\infty$) reduces MAE by 10.9\% and 32.9\% in AP and AR, respectively, yielding the best overall performance. The more pronounced improvement in AR indicates that traffic evolution in dense airspace operations is influenced not only by aircraft currently under control, but also by surrounding traffic beyond the controlled boundary. These results support the importance of modeling a broader airspace scope for accurate flow prediction.}

\subsection{Case study: online forecasting in operational deployment}\label{sec:case_study}

This section presents the real-world online forecasting performance of AeroSense in a production environment. \bwang{AeroSense is currently deployed in an operational streaming environment
as a supplementary demand-monitoring component rather than an autonomous
control system. Its forecasts are intended to be reviewed together with
surveillance displays, flight-plan information, and coordination records,
while all operational decisions and control instructions remain the
responsibility of qualified personnel. This human-in-the-loop positioning
is consistent with current ATM guidance on trustworthy human-centric
automation~\citep{eurocontrol2020flyai}.} During deployment, AeroSense continuously receives streaming ADS-B data from an Apache Kafka-based system \citep{kreps2011kafka} and generates 15-minute-ahead traffic forecasts at one-minute intervals. We emphasize that AeroSense was developed exclusively using data collected between March 1 and October 31, 2025, and subsequently deployed without any retraining or fine-tuning on the year 2026 (see potential improvement in Discussion \ref{ref:dis}). \textit{Consequently, any operational data from 2026 are entirely unseen by the model, making this a strict real-world out-of-time-domain evaluation.} \aqthenbw{For comparison, representative baselines, including GTR, MTEFP, and DFP-FG, are evaluated on the same unseen operational case study data.} The case study presented in this section is based on operational data collected on June 4, 2026, and is analyzed from the following four perspectives. 

\begin{figure*}[!hbt]
\centering
\includegraphics[width=\textwidth]{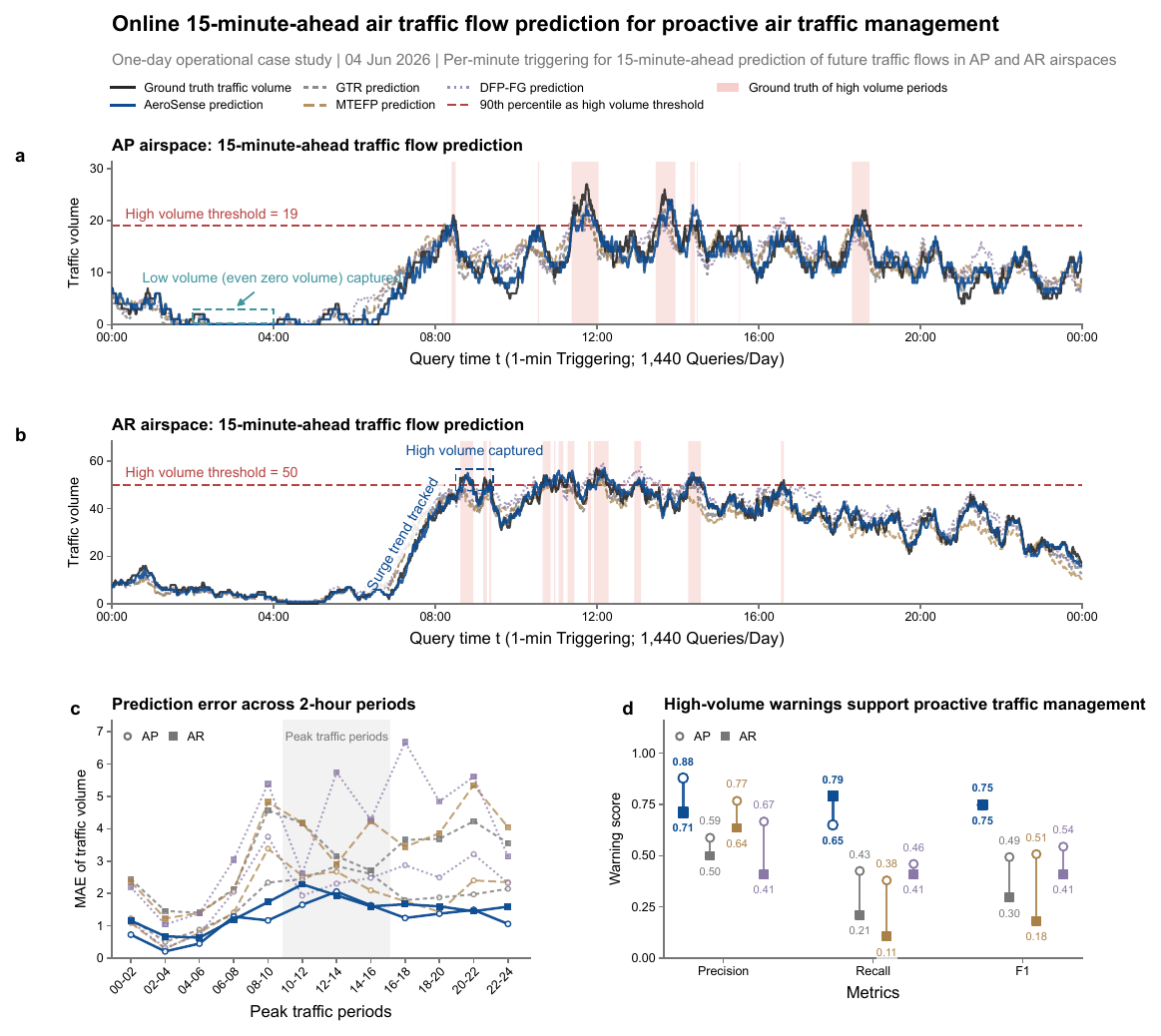}
\caption{
\textbf{One-day online operational evaluation of AeroSense.}
The model is deployed in a production environment and triggered once per minute, yielding 1,440 forecasting queries over a full operational day. Each query predicts the traffic volume in the AP and AR airspaces over the future interval $(t,t+15,\mathrm{min}]$. 
\aqthenbw{(a,b) Predicted traffic volumes from AeroSense, GTR, MTEFP, and DFP-FG together with the ground truth traffic volumes in the AP and AR airspaces, respectively.}
\aqthenbw{(c) Prediction MAE error across two-hour periods for AeroSense and the comparison baselines, where each point represents the average MAE of all per-minute forecasts within the corresponding time interval.}
\aqthenbw{(d) High-volume warning performance of AeroSense and the comparison baselines obtained by thresholding the predicted traffic volume, reported in terms of Precision, Recall, and F1 score.}
The results demonstrate that AeroSense accurately captures low-volume conditions, traffic surges, and sustained high-volume operations, while providing effective support for proactive traffic management.
}

\label{fig:online_case_study}
\end{figure*}

{\textbf{(1) Real-time tactical forecasting at arbitrary query times.}
\textit{A notable advantage of AeroSense is its ability to support real-time tactical forecasting at arbitrary query times.} Because AeroSense directly maps the instantaneous aircraft-state set to future traffic flow, it does not depend on a fixed historical \textit{look-back window (see also Fig. \ref{fig:motivation})}. As a result, forecasts can be generated whenever required by operational decision makers—every minute, every five minutes, or at any temporal resolution supported by the ADS-B system—without retraining the model. In the online case study shown in Fig.~\ref{fig:online_case_study}, AeroSense forecasting is triggered once per minute. For each query time $t$, the model takes the current airspace situation as input and predicts the traffic volume over the future interval $(t,t+15\mathrm{min}]$. Consequently, the horizontal axes in Fig.~\ref{fig:online_case_study}(a) and (b) represent consecutive query times throughout a 24-hour operational period, yielding a total of 1,440 prediction queries. This capability highlights a practical advantage of the proposed state-to-flow paradigm for streaming online forecasting.}

\textbf{(2) Accurate tracking of zero-flow periods, traffic surges, and high-volume operations without apparent temporal lag.} \aqthenbw{Compared with GTR, MTEFP, and DFP-FG, AeroSense more closely follows the ground truth traffic evolution, particularly during rapid traffic increases and sustained high-volume periods in both the AP and AR airspaces.} As shown in Fig.~\ref{fig:online_case_study}(a), during the early-morning low-volume period, the predicted flow accurately captures zero traffic conditions. As shown in Fig.~\ref{fig:online_case_study}(b), during the morning traffic surge, AeroSense successfully anticipates the rapid increase in future traffic demand without exhibiting an apparent temporal lag. Furthermore, the high-volume period highlighted by the \colorbox[HTML]{FAE5E2}{red shaded region} is also well predicted.

\binwang{\textbf{(3) Period-wise two-hour evaluation over one day.} To assess forecasting performance under different traffic conditions throughout the day, we partition the 24-hour operational period into twelve non-overlapping two-hour intervals and evaluate the prediction error within each interval. For each airspace $\mathcal{A}\in\{\mathrm{AP},\mathrm{AR}\}$, the period-wise MAE is computed by averaging all per-minute prediction errors whose query time $t$ falls within the corresponding interval. \aqthenbw{Overall, AeroSense consistently achieves lower period-wise MAE than GTR, MTEFP, and DFP-FG across the operational day, with particularly clear advantages during high-demand periods.} As shown in Fig.~\ref{fig:online_case_study}(c), AeroSense maintains consistently stable prediction accuracy throughout the day. The MAE ranges from 0.2 to 2.1 in the AP airspace and from 0.6 to 2.3 in the AR airspace, with average values of 1.20 and 1.46, respectively. The relatively larger errors occur during high-demand daytime periods, notably 12:00--14:00 in AP and 10:00--12:00 in AR, where traffic density and operational complexity are highest.}

\textbf{(4) High-volume early warning using the 90th-percentile threshold.}
Following the recommendation of ATM practitioners, \aqthenbw{we define the high-volume warning threshold as the 90th percentile of ground truth traffic volumes observed during the preceding week (19 for AP and 50 for AR). Thus, the threshold remains fully be available and is fixed before the evaluation day. A warning is issued when the predicted 15-minute-ahead traffic volume exceeds the corresponding threshold, and performance is evaluated as a binary classification task. This simple yet effective strategy avoids both training a separate classifier and using information from the evaluation day, thereby better reflecting prospective operation.} \aqthenbw{Compared with GTR, MTEFP, and DFP-FG, AeroSense achieves the strongest overall warning performance in both airspaces.} As shown in Fig.~\ref{fig:online_case_study}(d), AeroSense achieves Precision/Recall/F1 scores of 0.88/0.65/0.75 in AP and 0.71/0.79/0.75 in AR.

\bwang{Operationally, a predicted threshold exceedance would prompt further assessment using surveillance tracks, sector status, and coordination information, while all decisions remain with qualified personnel~\citep{eurocontrol2020flyai}. Because forecasts are refreshed every minute, operators can also assess whether alerts persist across consecutive updates and use subsequent observations to recalibrate local thresholds.} These results show that AeroSense can provide not only accurate traffic-volume forecasts but also \bwang{a technically feasible high-volume warning signal for human-in-the-loop operational assessment without additional retraining}.

\aqthenbw{For real-time online inference, the trained AeroSense model runs on an NVIDIA GeForce RTX 4090 with FP32 precision. It contains only 0.12M parameters and requires approximately 0.0128 GFLOPs per forecasting query. The average forward latency is $0.868 \pm 0.009$ ms, with a peak GPU memory usage of 10.71 MB, supporting lightweight and continuous online operation.}

\section{Discussion}
\textbf{Conclusions.} This paper reformulates air traffic flow prediction from a different perspective. We argue that the conventional macroscopic paradigm, which relies on aggregated time-series observations, introduces an inherent mismatch between the learning representation and the true physical state of air traffic. In practice, traffic in the TA is more naturally characterized as a dynamic, variable-cardinality set of aircraft evolving in continuous airspace. To address this mismatch, we propose \textit{AeroSense}, a state-to-flow modeling framework that directly maps microscopic aircraft states to macroscopic traffic flows. By explicitly representing the instantaneous airspace situation as a dynamic set, AeroSense removes the need for historical look-back windows and mitigates the information loss caused by spatial and temporal aggregation. Extensive experiments on a large-scale real-world TA dataset demonstrate that AeroSense consistently outperforms strong time-series baselines. The results show that directly modeling microscopic aircraft states leads to substantially higher predictive fidelity. Beyond improving average predictive accuracy, AeroSense also exhibits strong robustness during peak periods and nearly achieves Pareto-optimal performance under dayparting multi-objective  evaluation, underscoring its practical value for proactive ATM.

From a deployment perspective, \bwang{although the present evaluation is limited to one operational airport, the state-to-flow formulation is not tied to a specific airport or runway layout. AP and AR are local instantiations that can be redefined for other terminal airspaces or finer-grained sectors. Likewise, surrounding airspace coverage is a configurable deployment setting rather than a prerequisite for complete surveillance; the observable scope can be adapted to local ATM infrastructure, neighboring-airspace accessibility and data sharing policies.} AeroSense also simplifies real-time streaming inference. Because AeroSense does not require long historical windows of aggregated traffic statistics for online prediction, it can reduce storage overhead and simplify variable maintenance in streaming systems such as Apache Kafka \citep{kreps2011kafka}. {This property also aligns with recent efforts toward lightweight and deployable 4D trajectory prediction for real-time ATM applications \citep{tang2025trajectory}.}

\textbf{Limitations and future works.}\label{ref:dis} \bwang{The short-term aircraft traffic flow prediction represents only one dimension
of terminal airspace demand and should not be interpreted as a complete
surrogate for controller workload, complexity, conflict risk, capacity
utilization, or delay. These quantities capture complementary operational
dimensions and generally require additional information such as sector
configuration, traffic geometry, and controller procedures. Importantly, the proposed state-to-flow formulation is not tied to the AP/AR
spatial granularity. It can be naturally extended to finer-grained sectors by
redefining the target-region boundaries ~\citep{zhang2025short}. The present AP/AR experiments therefore provide a regional-level validation of the formulation rather than restricting
its applicability to these two airspaces.} 

This study primarily focuses on short-term 15-minute-ahead forecasting, as this horizon is among the most operationally relevant for ATC authorities. Accordingly, AeroSense is currently deployed for 15-minute-ahead online air traffic prediction in practice (see case study in Section \ref{sec:case_study}). Extending the framework to longer forecasting horizons, such as 30-minute-ahead and 90-minute-ahead prediction tasks~\citep{zhang2025short}, is both natural and an important direction for future research. In addition, the current framework primarily models aircraft-state-driven traffic dynamics using a static learning paradigm. \binwang{Although AeroSense was trained on historical data from 2025 and directly deployed on unseen 2026 operational data without fine-tuning, it still captured the major traffic evolution patterns, suggesting promising temporal transferability of the proposed state-to-flow paradigm. \bwang{The current ADS-B formulation does not explicitly encode route membership, runway assignments, metering fixes, or controller-imposed sequencing. These operational relations may provide complementary information and could be incorporated through multimodal state representations when consistently synchronized data become available.} Future work will explore dynamic online learning and adaptation mechanisms \citep{chen2026uncertainty, li2026text} to continuously update the model with newly accumulated operational data and improve performance under evolving traffic conditions. Furthermore, integrating external factors, such as weather conditions~\citep{zeng2024improved} and richer air traffic control instruction information~\citep{guo2024integrating}, may further enhance predictive performance and robustness under rapidly changing operational environments.} 
\section{Methods}
\subsection{Aircraft-level state representation}
\label{sec:physics_features}

We next construct an aircraft-state representation from trajectory data to encode aircraft kinematics, partial control-intent cues, boundary interactions, and temporal context. Each aircraft at time $t$ is represented by a state vector $\mathbf{s}_i^{(t)} \in \mathbb{R}^{D_{\mathrm{in}}}$, where $D_{\mathrm{in}}=18$ in this study. The state vector is formed by concatenating the following five groups of states derived from raw ADS-B data:

\noindent\textbf{1. State of aircraft location  $\mathbf{f}_{l}$.} 
The location state describes the aircraft position, including latitude $\varphi$, longitude $\lambda$, and barometric altitude $H$:
\begin{equation}
\label{formula_start}
\mathbf{f}_{l} = [\varphi, \lambda, H] \in \mathbb{R}^3 .
\end{equation}
\noindent\textbf{2. State of aircraft kinematic $\mathbf{f}_{k}$.}
The kinematic state captures instantaneous motion using ground speed $v_{gs}$, vertical speed $v_{vs}$, and  heading angle $\theta$:
\begin{equation}
    \quad
    \mathbf{f}_{k} = [v_{gs}, v_{vs} , \theta] \in \mathbb{R}^3 .
\end{equation}
\noindent\textbf{3. State of \bwang{partial control-intent cues} $\mathbf{f}_{c}$.} \bwang{We include the dialed airspeed $v_{dial}$  and dialed altitude from the Mode Control Panel as partial cues of the aircraft’s near-term control intent:}
\begin{equation}
    \mathbf{f}_{c} = [v_{dial}, h_{dial}] \in \mathbb{R}^2.
\end{equation}
\bwang{These variables do not represent complete ATC clearances or operational intent.}

\noindent\textbf{4. State of boundary interactions $\mathbf{f}_{b}$.} 
To explicitly encode how an aircraft relates to the surrounding airspace geometry, we introduce boundary-interaction state that describe both its spatial position relative to a target airspace and its motion trend with respect to that airspace. Two quantities are defined.

\textit{1) State of  boundary proximity $d_{\mathcal{A}}$:}
To capture geometric effects associated with airspace entry and exit, we compute the minimum distance from the aircraft location $\mathbf{p}_i$ to the boundary $\partial \Omega_{\mathrm{ctr}}$ of airspace $\mathcal{A}$, i.e., AR or AP:

\begin{equation}
d_{\mathcal{A}} = \min_{\mathbf{b} \in  \mathcal{A}} \operatorname{dist}(p_i, b).
\end{equation}
This quantity reflects how close the aircraft is to the corresponding airspace boundary and therefore provides a direct cue about potential near-term transitions across airspaces.

\textit{2) Approach factor $\alpha_{\mathcal{A}}$:} To distinguish approaching traffic from bypassing or departing traffic, we further characterize the aircraft motion trend relative to the airspace center $\mathbf{c}_{\mathcal{A}}$. Let $\mathbf{r}_c = \mathbf{c}_{\mathcal{A}} - \mathbf{p}_i$ denote the relative position vector from the aircraft to the airspace center. The approach factor is defined as the cosine similarity between the velocity vector $\mathbf{v}_i$ and $\mathbf{r}_c$:

\begin{equation}
    \alpha_{\mathcal{A}} = \frac{\mathbf{v}_i \cdot \mathbf{r}_{c}}{\|\mathbf{v}_i\|_2 \|\mathbf{r}_{c}\|_2 + \epsilon}
\end{equation}
Here, $\alpha_{\mathcal{A}} > 0$ indicates motion toward the airspace center, whereas $\alpha_{\mathcal{A}} < 0$ indicates motion away from it.

We compute these states for both AP and AR, yielding:
\begin{equation}
    \mathbf{f}_{b} = [d_{AP}, d_{AR}, \alpha_{AP}, \alpha_{AR}, I_{AP}, I_{AR}] \in \mathbb{R}^6,
\end{equation}
where $I_{\mathcal{A}} \in \{0,1\}$ denotes the airspace inclusion indicator for aircraft $i$, with $I_{\mathcal{A}} = 1$ if the aircraft is currently located within airspace $\mathcal{A}$ and $0$ otherwise. Together, these states provide complementary geometric and kinematic cues for describing how aircraft interact with the AP and AR boundaries.

\noindent\textbf{5. State of temporal context $\mathbf{f}_{t}$.}
Temporal context is encoded using cyclical embeddings to capture recurring temporal patterns in air traffic, such as intraday and intrahour variability. Given the timestamp at time $t$, we extract the hour $h_t \in [0, 24)$ and minute $m_t \in [0, 60)$. The temporal feature vector is then defined as


\begin{equation}
\begin{aligned}
\mathbf{f}_{t} = \bigl[&
\sin\left(\frac{2\pi h_t}{24}\right),
\cos\left(\frac{2\pi h_t}{24}\right),\\
&
\sin\left(\frac{2\pi m_t}{60}\right),
\cos\left(\frac{2\pi m_t}{60}\right)
\bigr] \in \mathbb{R}^4
\end{aligned}
\end{equation}
where the periods 24 and 60 correspond to the cycles used to normalize the hour and minute components, respectively.

The final aircraft state is constructed via concatenation ($\oplus$) as 
\begin{equation}
\mathbf{s}_i^{(t)} =
\mathrm{Norm}\!\left(\mathbf{f}_{l} \oplus \mathbf{f}_{k} \oplus \mathbf{f}_{c}\right)
\oplus \mathbf{f}_{b} \oplus \mathbf{f}_{t}
\in \mathbb{R}^{18}
\end{equation}
where $\mathrm{Norm}(\cdot)$ applies z-score normalization to $\mathbf{f}_{l}$, $\mathbf{f}_{k}$, and $\mathbf{f}_{c}$, while $\mathbf{f}_{b}$ and $\mathbf{f}_{t}$ are left unnormalized.

\subsection{Model architecture of AeroSense}
\label{sec:model_arch}

This subsection presents the AeroSense architecture, with particular emphasis on variable-cardinality set handling, inter-aircraft interaction modeling, and task-specific prediction.

\begin{figure*}[!ht]
  \centering
  \includegraphics[width=\linewidth]{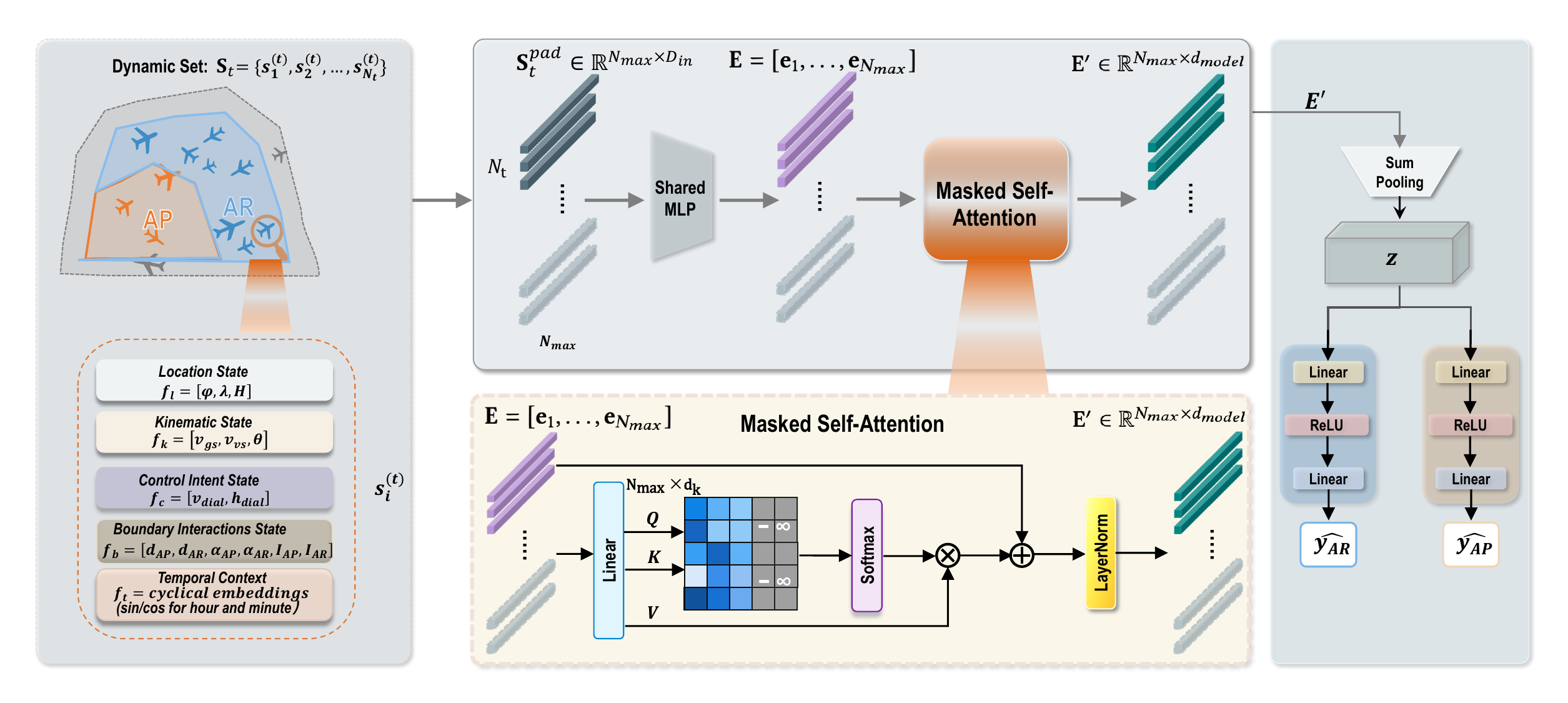} 
  \vspace{-1em}
  \caption{\textbf{The architecture of the proposed AeroSense model.}
This figure illustrates the key variables and computational steps of the inference pipeline, corresponding to Equations~\ref{formula_start}–\ref{formula_end}.
The instantaneous airspace situation at time $t$ is represented as a dynamic set of aircraft states 
$\mathbf{S}_t=\{\mathbf{s}_1^{(t)},\mathbf{s}_2^{(t)},\dots,\mathbf{s}_{N_t}^{(t)}\}$, 
where each state encodes aircraft location, kinematics, control-intent cues, boundary interactions, and temporal context. 
\aqthenbw{For batched computation, the variable-cardinality aircraft sets are padded to the largest set cardinality within the current mini-batch and encoded through a shared MLP, followed by masked self-attention to model inter-aircraft interactions \aqthenbw{while masking padded states in the key and value positions.}
}
The resulting aircraft representations are aggregated via \textit{SumPooling} to produce a global airspace-level representation, which is then fed into two decoupled prediction branches for future AP and AR traffic flow prediction.}
  \label{fig:framework}
\end{figure*}

\subsubsection{Variable-cardinality set handling via container initialization}
The input to AeroSense is the aircraft state set $\mathbf{S}_t$ defined in Definition~\ref{def:as}. 
\aqthenbw{During batched training and evaluation, each sample $\mathbf{S}_t$ is first padded to $N_{max}$ and represented by a padded matrix $\mathbf{S}_t^{\mathrm{pad}} \in \mathbb{R}^{N_{max} \times D_{in}}$:}

\begin{equation}
    \aqthenbw{\mathbf{S}_t^{\mathrm{pad}} =
    [\mathbf{s}_1^{(t)}, \dots, \mathbf{s}_{N_t}^{(t)},
    \underbrace{\mathbf{0}, \dots, \mathbf{0}}_{N_{max}-N_t}]^\top}
\end{equation}
\aqthenbw{\aqthenbw{For a mini-batch $\mathcal{B}$ containing $K$ samples, we denote
\begin{equation}
    \mathcal{B}
    \overset{\mathrm{def}}{=}
    \left\{
    \mathbf{S}_{t_k}
    \right\}_{k=1}^{K},
\end{equation}
where $t_k$ denotes the query time of the $k$-th sample. Here, $N_{max}=\max_{1\leq k\leq K}N_{t_k}$ denotes the largest number of aircraft among the samples in the current mini-batch. For a sample at time $t$,}
the first $N_t$ rows correspond to the valid aircraft states in $\mathbf{S}_t^{\mathrm{pad}}$, while the remaining $N_{max}-N_t$ rows are zero-padding vectors. The resulting matrices in the same mini-batch therefore have the same size and can be processed in parallel. During online inference for an individual traffic situation, the input is constructed directly from the $N_t$ observed aircraft states, and no padding is required.}

\subsubsection{Deep representation learning from input state}
Given the padded matrix $\mathbf{S}_t^{\mathrm{pad}}$,
to learn aircraft-level deep representations, we use a weight-shared MLP to project these physical states into a latent space. 
\aqthenbw{Let $\mathbf{H}^{(0)}=\mathbf{S}_t^{\mathrm{pad}}$ denote the input to the first layer}, the propagation at the $l$-th layer is defined as

\begin{equation} \mathbf{H}^{(l)} = \mathrm{Dropout}\left( \mathrm{ReLU}\left( \mathrm{BN}\left( \mathbf{H}^{(l-1)}\mathbf{W}^{(l)\top} + \mathbf{b}^{(l)} \right) \right) \right). \end{equation}
where $\mathbf{W}^{(l)}$ and $\mathbf{b}^{(l)}$ are the learnable parameters of the $l$-th layer. Here, $\text{BN}(\cdot)$ and $\text{Dropout}(\cdot)$ denote batch normalization~\citep{ioffe2015batch} and dropout regularization~\citep{srivastava2014dropout}, respectively, and $\mathrm{ReLU}(\cdot)$ denotes the rectified linear unit activation function. The resulting aircraft embedding is denoted by 

\begin{equation}\label{ref:e}
\mathbf{E}
=
\mathbf{H}^{(L)}
=
[\mathbf{e}_1,\ldots,\mathbf{e}_{N_{max}}]^\top
\in
\mathbb{R}^{N_{max}\times d_{model}},
\end{equation}
where $L$ denotes the number of fully connected layers, and two layers are employed in the study. Here, $d_{model}$ is the latent space dimension, and $\mathbf{e}_i$ denotes the representation of the $i$-th aircraft.

\subsubsection{Aircraft interaction modeling via self-attention}
{
To capture implicit inter-aircraft dependencies and global correlations within the traffic scene, we employ a multi-head self-attention mechanism \citep{vaswani2017attention}. }

{To prevent these artificial padding states from interfering with the interaction modeling, we introduce an attention masking mechanism. 

\aqthenbw{The mask prevents padded aircraft states from contributing as keys and values in the self-attention operation. }
We define a mask matrix $M \in \mathbb{R}^{N_{max} \times N_{max}}$ as follows:

\begin{equation}
    \aqthenbw{M_{ij} =
    \begin{cases}
    0, & \text{if } 1 \le j \le N_t, \\
    -\infty, & \text{if } N_t < j \le N_{max},
    \end{cases}
    \qquad 1 \le i \le N_{max}.}
\end{equation}
}

{Note $\mathbf{E} = [\mathbf{e}_1, \dots, \mathbf{e}_{N_{max}}]^\top$ denote the matrix of aircraft embeddings, for the $j$-th attention head, the query ($\mathbf{Q}_j$), key ($\mathbf{K}_j$), and value ($\mathbf{V}_j$) matrices are computed via linear projections:
\begin{equation}
    \mathbf{Q}_j = \mathbf{E}\mathbf{W}_Q^j, \quad 
    \mathbf{K}_j = \mathbf{E}\mathbf{W}_K^j, \quad 
    \mathbf{V}_j = \mathbf{E}\mathbf{W}_V^j.
\end{equation}}

{Then, the scaled dot-product $\mathbf{Q}_j \mathbf{K}_j^\top$ computes the pairwise similarity as the interaction strength between different aircraft. The mask matrix $M$ is then added to this result before applying the Softmax function to obtain the normalized attention weights:
\begin{equation}
    \mathbf{A}_j = \text{Softmax}\left(\frac{\mathbf{Q}_j \mathbf{K}_j^\top}{\sqrt{d_k}} + M\right) \mathbf{V}_j
\end{equation}}

{By assigning $-\infty$ to the padding positions in $M$, their corresponding attention weights are strictly driven to $0$ after the Softmax operation. Consequently, the padded items are completely filtered out, allowing the model to focus solely on the structural relationships and interactions among the truly valid aircraft.}

\aqthenbw{The outputs of all attention heads are concatenated and projected, and then combined with the original embedding through a residual connection and layer normalization:
\begin{equation}\label{ref:eprime}
    \mathbf{E}'
    =
    \mathrm{LayerNorm}\left(
    \mathbf{E}
    +
    \mathrm{Concat}\left(
    \mathbf{A}_1,\ldots,\mathbf{A}_H
    \right)\mathbf{W}_O
    \right),
\end{equation}
where $\mathbf{W}_O \in \mathbb{R}^{H d_k \times d_{model}}$ is a learnable projection matrix, and $H$ denotes the number of attention heads. We denote
$\mathbf{E}' = [\mathbf{e}'_1,\ldots,\mathbf{e}'_{N_{max}}]^\top$,
where $\mathbf{e}'_i$ is the updated representation of aircraft $i$ after self-attention.}

\subsubsection{Padding state filtering and summation pooling}

We next employ a permutation-invariant aggregation function $\rho$ to obtain a fixed-length global context vector $\mathbf{z}$. Unlike classification tasks, where Max or Mean pooling is often sufficient, traffic flow prediction is  a volume regression problem that depends on the number of aircraft in the set. Motivated by deep sets theory \citep{zaheer2017deep}, we therefore instantiate $\rho$ as element-wise summation (hereafter referred to as \textit{SumPooling}), which preserves traffic scale in the global representation. Because $\mathbf{S}_t^{\mathrm{pad}}$ contains $N_{max}-N_t$ padded states, we introduce an indicator function $\mathbb{I}_{filter}(i)$ to remove them during aggregation:

\begin{equation}
\mathbb{I}_{filter}(i) = \begin{cases} 
1 & \text{if } 1 \le i \le N_t \\ 
0 & \text{otherwise} 
\end{cases}
\end{equation}

We formalize this \textit{SumPooling} operation to compute the global representation $\mathbf{z}$ as follows:
\begin{equation}
\label{eq:pooling}
\mathbf{z} = \rho(\{\mathbf{e}'_1, \dots, \mathbf{e}'_{N_{max}}\}) = \sum_{i=1}^{N_{max}} \mathbf{e}'_i \cdot \mathbb{I}_{filter}(i)
\end{equation}
By explicitly summing the valid states, this \textit{SumPooling} captures total traffic accumulation, ensuring that the model remains sensitive to aircraft count rather than only to average state characteristics.

\subsubsection{Prediction via decoupled heads}
Because the AP and AR exhibit distinct traffic patterns and flow distributions, we design two independent decoder prediction heads $g_{\mathcal{A}}(\cdot)$, one for each airspace $\mathcal{A}$. The global context vector $\mathbf{z}$ is fed into both branches:

\begin{equation}
\label{formula_end}
\hat{y}_{\mathcal{A}} = g_{\mathcal{A}}(\mathbf{z}) = \mathbf{W}_{\mathcal{A}}^{(2)} \mathrm{ReLU} \left( \mathbf{W}_{\mathcal{A}}^{(1)} \mathbf{z} + \mathbf{b}_{\mathcal{A}}^{(1)} \right) + b_{\mathcal{A}}^{(2)}
\end{equation}
where $\mathbf{W}_{\mathcal{A}}^{(l)}$ and $\mathbf{b}_{\mathcal{A}}^{(l)}$ denote the learnable weight matrix and bias vector of the $l$-th layer for airspace $\mathcal{A}$. The final model output is $\hat{\mathbf{Y}} = [\hat{y}_{AP}, \hat{y}_{AR}]^\top$.

\subsubsection{Computational complexity}

\aqthenbw{For a traffic situation containing $N_t$ aircraft, the shared aircraft encoder and SumPooling scale linearly with $N_t$, while the decoupled prediction heads operate on the fixed-dimensional global representation and are independent of aircraft cardinality. The self-attention module evaluates pairwise relations among aircraft and has quadratic complexity, $O(N_t^2)$. Therefore, with fixed feature and hidden dimensions, the overall computational complexity of AeroSense with respect to aircraft cardinality is $O(N_t^2)$.}

\subsection{Multi-task loss function}
We adopt the Huber loss as the training objective. This loss combines the smooth optimization behavior of MSE for small errors with the robustness of MAE to large deviations. The total loss is defined as the sum of prediction errors over both target airspaces:

\begin{equation}
\mathcal{L}(\Theta) = \frac{1}{B} \sum_{k=1}^{B} \sum_{\mathcal{A} \in \{AP, AR\}} \mathcal{L}_{\delta}(y_{\mathcal{A}}^{(k)} - \hat{y}_{\mathcal{A}}^{(k)})
\end{equation}
where $B$ denotes the batch size. The Huber loss is defined as follows, with $\delta = 1.0$ in our experiments and $a = y - \hat{y}$ denoting the prediction residual:

\begin{equation}
\mathcal{L}_{\delta}(a) = \left\{ \begin{array}{ll} \frac{1}{2}a^2 & \text{if } |a| \le \delta, \\ \delta (|a| - \frac{1}{2}\delta) & \text{otherwise.} \end{array} \right.
\end{equation}

\section{Data availability}
Source data will be provided with this paper.

\section{Code availability}
The source code is publicly available at \url{https://github.com/aspjy/aerosense_open}.


\begin{thebibliography}{58}
\providecommand{\natexlab}[1]{#1}
\providecommand{\url}[1]{{#1}}
\providecommand{\urlprefix}{URL }
\providecommand{\doi}[1]{\url{https://doi.org/#1}}
\providecommand{\eprint}[2][]{\url{#2}}
 \bibcommenthead

\bibitem[{Bergmeir and Ben{\'i}tez(2012)}]{bergmeir2012crossvalidation}
Bergmeir C, Ben{\'i}tez JM (2012) On the use of cross-validation for time series predictor evaluation. Information Sciences 191:192--213. \doi{10.1016/j.ins.2011.12.028}

\bibitem[{Brooker(2008)}]{brooker2008sesar}
Brooker P (2008) {SESAR} and nextgen: Investing in new paradigms. Journal of Navigation 61(2):195–208

\bibitem[{Cao et~al.(2026)Cao, Dai, Han, and Xiong}]{ICLR2026_7a3284ce}
Cao F, Dai L, Han J, et~al (2026) Enhancing multivariate time series forecasting with global temporal retrieval. In: Vondrick C, Hariharan B, Raffel C, et~al (eds) International Conference on Learning Representations, pp 75430--75458, \urlprefix\url{https://proceedings.iclr.cc/paper_files/paper/2026/file/7a3284ce53f620d244fca0c79d94c478-Paper-Conference.pdf}

\bibitem[{Chen et~al.(2016)Chen, Hu, Ma, and Yin}]{chen2016network}
Chen D, Hu M, Ma Y, et~al (2016) A network-based dynamic air traffic flow model for short-term en route traffic prediction. Journal of Advanced Transportation 50(8):2174--2192

\bibitem[{Chen et~al.(2026)Chen, Luo, Huang, Jiang, Shi, Zhang, and Gao}]{chen2026uncertainty}
Chen J, Luo Y, Huang X, et~al (2026) Uncertainty-aware online time series multi-step forecasting framework in cloud systems. IEEE Transactions on Knowledge and Data Engineering

\bibitem[{Dalichampt and Plusquellec(2007)}]{dalichampt2007hourly}
Dalichampt M, Plusquellec C (2007) Hourly entry count versus occupancy count relationship: Definitions and indicators (i). Tech. Rep. EEC Note No. 15/07, EUROCONTROL Experimental Centre

\bibitem[{Du et~al.(2024)Du, Chen, Li, Cao, and Lv}]{du2024spatial}
Du W, Chen S, Li Z, et~al (2024) A spatial-temporal approach for multi-airport traffic flow prediction through causality graphs. IEEE Transactions on Intelligent Transportation Systems 25(1):532--544

\bibitem[{{EUROCONTROL}(2024)}]{eurocontrol2024flowconops}
{EUROCONTROL} (2024) Flow conops. Tech. rep., EUROCONTROL Network Management Directorate

\bibitem[{{European Aviation Artificial Intelligence High Level Group}(2020)}]{eurocontrol2020flyai}
{European Aviation Artificial Intelligence High Level Group} (2020) The {FLY AI} report: Demystifying and accelerating {AI} in aviation/{ATM}. Tech. rep., EUROCONTROL, \urlprefix\url{https://www.eurocontrol.int/sites/default/files/2020-03/eurocontrol-fly-ai-report-032020.pdf}

\bibitem[{Gui et~al.(2020)Gui, Zhou, Wang, Liu, and Sun}]{gui2020machine}
Gui G, Zhou Z, Wang J, et~al (2020) Machine learning aided air traffic flow analysis based on aviation big data. IEEE Transactions on Vehicular Technology 69(5):4817--4826

\bibitem[{Guo et~al.(2024{\natexlab{a}})Guo, Zhang, Yan, Zhang, and Lin}]{Guo2024FlightBERTpp}
Guo D, Zhang Z, Yan Z, et~al (2024{\natexlab{a}}) {FlightBERT}++: A non-autoregressive multi-horizon flight trajectory prediction framework. Proceedings of the AAAI Conference on Artificial Intelligence 38(1):127–134

\bibitem[{Guo et~al.(2024{\natexlab{b}})Guo, Zhang, Yang, Zhang, Yang, and Lin}]{guo2024integrating}
Guo D, Zhang Z, Yang B, et~al (2024{\natexlab{b}}) Integrating spoken instructions into flight trajectory prediction to optimize automation in air traffic control. Nature Communications 15(1):9662

\bibitem[{Hewamalage et~al.(2023)Hewamalage, Ackermann, and Bergmeir}]{hewamalage2023forecast}
Hewamalage H, Ackermann K, Bergmeir C (2023) Forecast evaluation for data scientists: common pitfalls and best practices. Data Mining and Knowledge Discovery 37:788--832. \doi{10.1007/s10618-022-00894-5}

\bibitem[{{International Civil Aviation Organization}(2018)}]{icao2018annex11}
{International Civil Aviation Organization} (2018) Annex 11 to the Convention on International Civil Aviation: Air Traffic Services, 15th edn. International Civil Aviation Organization, Montreal, Canada

\bibitem[{{International Civil Aviation Organization}(2025)}]{icao2025atmas}
{International Civil Aviation Organization} (2025) Air traffic management automation system implementation and operations guidance document. Guidance Document Edition 1.5, International Civil Aviation Organization, \urlprefix\url{https://www.icao.int/sites/default/files/APAC/Documents/ATMAS-Implementation-and-Operations-Guidance-Document-Edition-1.5.pdf}, asia and Pacific Office

\bibitem[{Ioffe and Szegedy(2015)}]{ioffe2015batch}
Ioffe S, Szegedy C (2015) Batch normalization: Accelerating deep network training by reducing internal covariate shift. In: International Conference on Machine Learning (ICML), pp 448--456

\bibitem[{Jiang and Zhang(2024)}]{jiang2024characteristics}
Jiang F, Zhang Z (2024) Characteristics of air traffic flow in terminal airspace: A multiplex recurrence network analysis. IEEE Transactions on Intelligent Transportation Systems 25(10):14803--14815

\bibitem[{Jurinić et~al.(2024)Jurinić, Juričić, Antulov-Fantulin, and Samardžić}]{jurinic2024defining}
Jurinić T, Juričić B, Antulov-Fantulin B, et~al (2024) Defining terminal airspace air traffic complexity indicators based on air traffic controller tasks. Aerospace 11(5)

\bibitem[{Kim et~al.(2026)Kim, Yoon, and Lee}]{kim2026probabilistic}
Kim K, Yoon S, Lee K (2026) Probabilistic multi-agent aircraft landing time prediction. In: AIAA SCITECH 2026 Forum, p 1764, \doi{10.2514/6.2026-1764}

\bibitem[{Kl{\"o}tergens et~al.(2023)Kl{\"o}tergens, Acevedo, Firmansyah, Antiqui, Madhusudhanan, and Jameel}]{klotergens2023predicting}
Kl{\"o}tergens C, Acevedo C, Firmansyah I, et~al (2023) Predicting deviation of flight entry into air sector using machine learning techniques. In: 2023 IEEE/AIAA 42nd Digital Avionics Systems Conference (DASC), IEEE, pp 1--10

\bibitem[{Kreps et~al.(2011)Kreps, Narkhede, Rao et~al.}]{kreps2011kafka}
Kreps J, Narkhede N, Rao J, et~al (2011) Kafka: A distributed messaging system for log processing. In: Proceedings of the NetDB, Athens, Greece, pp 1--7

\bibitem[{Lee et~al.(2019)Lee, Lee, Kim, Kosiorek, Choi, and Teh}]{lee2019set}
Lee J, Lee Y, Kim J, et~al (2019) Set transformer: A framework for attention-based permutation-invariant neural networks. In: International Conference on Machine Learning (ICML), pp 3744--3753

\bibitem[{Li et~al.(2026)Li, Wang, Yan, Zhang, Deng, Sun, Han, and Gong}]{li2026text}
Li L, Wang Y, Yan J, et~al (2026) From text to forecasts: Bridging modality gap with temporal evolution semantic space. arXiv preprint arXiv:260312664

\bibitem[{Li et~al.(2024)Li, Jiang, Liu, Li, and Zhou}]{li2024airspace}
Li Y, Jiang B, Liu W, et~al (2024) Airspace situation analysis of terminal area traffic flow prediction based on big data and machine learning methods. Big Data Research 35:100425

\bibitem[{Lin et~al.(2019)Lin, Zhang, and Liu}]{lin2019deep}
Lin Y, Zhang Jw, Liu H (2019) Deep learning based short-term air traffic flow prediction considering temporal–spatial correlation. Aerospace Science and Technology 93:105113

\bibitem[{Liu et~al.(2026)Liu, Hu, and Yang}]{LIU2026111029}
Liu H, Hu M, Yang L (2026) A directional flow pattern-driven framework for fine-grained air traffic flow prediction. Aerospace Science and Technology 168:111029. \doi{https://doi.org/10.1016/j.ast.2025.111029}, \urlprefix\url{https://www.sciencedirect.com/science/article/pii/S1270963825010922}

\bibitem[{Liu et~al.(2024)Liu, Hu, Zhang, Wu, Wang, Ma, and Long}]{liu2024itransformer}
Liu Y, Hu T, Zhang H, et~al (2024) i{T}ransformer: Inverted transformers are effective for time series forecasting. In: International Conference on Learning Representations (ICLR)

\bibitem[{Ma et~al.(2024)Ma, Alam, Cai, and Delahaye}]{ma2024text}
Ma C, Alam S, Cai Q, et~al (2024) Text-{E}nriched air traffic flow modeling and prediction using transformers. IEEE Transactions on Intelligent Transportation Systems 25(7):7963--7976

\bibitem[{Nagaoka and Brown(2014)}]{nagaoka2014review}
Nagaoka S, Brown M (2014) A review of safety indices for trajectory-based operations in air traffic management. Transactions of the Japan Society for Aeronautical and Space Sciences, Aerospace Technology Japan 12(APISAT-2013):a43--a49

\bibitem[{Nguyen and Liem(2025)}]{nguyen2025multi}
Nguyen CHC, Liem RP (2025) Multi-aircraft attention-based model for perceptive arrival transit time prediction. Advanced Engineering Informatics 64:103067. \doi{10.1016/j.aei.2024.103067}

\bibitem[{Niu et~al.(2026)Niu, Deng, and Tong}]{niu2026phaseformer}
Niu Y, Deng J, Tong Y (2026) Phaseformer: From patches to phases for efficient and effective time series forecasting. In: International Conference on Learning Representations (ICLR)

\bibitem[{Olive and Basora(2020)}]{olive2020detection}
Olive X, Basora L (2020) Detection and identification of significant events in historical aircraft trajectory data. Transportation Research Part C: Emerging Technologies 119:102737. \doi{10.1016/j.trc.2020.102737}

\bibitem[{Olive and Morio(2019)}]{olive2019trajectory}
Olive X, Morio J (2019) Trajectory clustering of air traffic flows around airports. Aerospace Science and Technology 84:776--781. \doi{10.1016/j.ast.2018.11.031}

\bibitem[{Pang et~al.(2023)Pang, Hu, Lieber, Cooke, and Liu}]{pang2023workload}
Pang Y, Hu J, Lieber CS, et~al (2023) Air traffic controller workload level prediction using conformalized dynamical graph learning. Advanced Engineering Informatics 57:102113. \doi{10.1016/j.aei.2023.102113}

\bibitem[{Patrikar et~al.(2025)Patrikar, Dantas, Moon, Hamidi, Ghosh, Keetha, Higgins, Chandak, Yoneyama, and Scherer}]{patrikar2025image}
Patrikar J, Dantas J, Moon B, et~al (2025) Image, speech, and {ADS-B} trajectory datasets for terminal airspace operations. Scientific Data 12(1):468

\bibitem[{Rebollo and Balakrishnan(2014)}]{rebollo2014characterization}
Rebollo JJ, Balakrishnan H (2014) Characterization and prediction of air traffic delays. Transportation Research Part C: Emerging Technologies 44:231--241

\bibitem[{{SESAR 3 Joint Undertaking}(2023)}]{sesar2023solution53b}
{SESAR 3 Joint Undertaking} (2023) {SESAR Solution 53B SPR-INTEROP/OSED for V3: Part I}. Technical report, SESAR 3 Joint Undertaking, \urlprefix\url{https://www.sesarju.eu/sites/default/files/documents/solution/Sol%20Pj18%20W2%2053B%20SPR-INTEROP_OSED%20Part%20I%20V3.pdf}, pJ18-W2 4DSkyways

\bibitem[{Shi et~al.(2021)Shi, Xu, and Pan}]{shi2021ConstrainedLSTM}
Shi Z, Xu M, Pan Q (2021) {4-D} flight trajectory prediction with constrained lstm network. IEEE Transactions on Intelligent Transportation Systems 22(11):7242--7255

\bibitem[{Srivastava et~al.(2014)Srivastava, Hinton, Krizhevsky, Sutskever, and Salakhutdinov}]{srivastava2014dropout}
Srivastava N, Hinton G, Krizhevsky A, et~al (2014) Dropout: a simple way to prevent neural networks from overfitting. The Journal of Machine Learning Research 15(1):1929--1958

\bibitem[{Strohmeier et~al.(2014)Strohmeier, Schäfer, Lenders, and Martinovic}]{strohmeier2014realities}
Strohmeier M, Schäfer M, Lenders V, et~al (2014) Realities and challenges of nextgen air traffic management: the case of {ADS-B}. IEEE Communications Magazine 52(5):111--118

\bibitem[{Tang et~al.(2025)Tang, Dai, Huang, Hao, and Xie}]{tang2025trajectory}
Tang W, Dai J, Huang Z, et~al (2025) {4D} trajectory lightweight prediction algorithm based on knowledge distillation technique. Frontiers in Neurorobotics 19:1643919

\bibitem[{Vaswani et~al.(2017)Vaswani, Shazeer, Parmar, Uszkoreit, Jones, Gomez, Kaiser, and Polosukhin}]{vaswani2017attention}
Vaswani A, Shazeer N, Parmar N, et~al (2017) Attention is all you need. In: Advances in Neural Information Processing Systems (NeurIPS)

\bibitem[{Vos et~al.(2024)Vos, Sun, and Hoekstra}]{vos2024transformer}
Vos R, Sun J, Hoekstra J (2024) A transformer-based trajectory prediction model to support air traffic demand forecasting. In: Proceedings of the 11th International Conference for Research in Air Transportation

\bibitem[{Wandelt et~al.(2025)Wandelt, Chen, and Sun}]{wandelt2025flight}
Wandelt S, Chen X, Sun X (2025) Flight delay prediction: A dissecting review of recent studies using machine learning. IEEE Transactions on Intelligent Transportation Systems 26(4):4283--4297

\bibitem[{Wu et~al.(2021)Wu, Xu, Wang, and Long}]{wu2021autoformer}
Wu H, Xu J, Wang J, et~al (2021) Autoformer: Decomposition transformers with auto-correlation for long-term series forecasting. In: Advances in Neural Information Processing Systems (NeurIPS), pp 22419--22430

\bibitem[{Wu et~al.(2023)Wu, Hu, Liu, Zhou, Wang, and Long}]{wu2023timesnet}
Wu H, Hu T, Liu Y, et~al (2023) Times{N}et: Temporal 2{D}-variation modeling for general time series analysis. In: International Conference on Learning Representations (ICLR)

\bibitem[{Wu et~al.(2024)Wu, Yang, Chen, Lin, and Yang}]{wu2024long}
Wu Y, Yang J, Chen X, et~al (2024) Long-{T}erm airport network performance forecasting with linear diffusion graph networks. IEEE Transactions on Intelligent Transportation Systems 25(11):18264--18278

\bibitem[{Yan et~al.(2022)Yan, Yang, Li, and Lin}]{yan2022deep}
Yan Z, Yang H, Li F, et~al (2022) A deep learning approach for short-term airport traffic flow prediction. Aerospace 9(1)

\bibitem[{Yan et~al.(2023)Yan, Yang, Wu, and Lin}]{yan2023multi}
Yan Z, Yang H, Wu Y, et~al (2023) A multi-view attention-based spatial–temporal network for airport arrival flow prediction. Transportation Research Part E: Logistics and Transportation Review 170:102997

\bibitem[{Ye et~al.(2026)Ye, Guo, Du, Luo, and Lin}]{ye2026multi}
Ye F, Guo D, Du H, et~al (2026) Multi-resolution temporal encoding enhanced air traffic flow prediction with frequency-aware decomposition. Transportation Research Part E: Logistics and Transportation Review 213:104976

\bibitem[{Yin et~al.(2025)Yin, Zhang, Zhang, Zhang, and Xiang}]{yin2025aircraft}
Yin Y, Zhang S, Zhang Y, et~al (2025) Aircraft trajectory prediction in terminal airspace with intentions derived from local history. Neurocomputing 615:128843

\bibitem[{Zaheer et~al.(2017)Zaheer, Kottur, Ravanbakhsh, Poczos, Salakhutdinov, and Smola}]{zaheer2017deep}
Zaheer M, Kottur S, Ravanbakhsh S, et~al (2017) Deep sets. In: Advances in Neural Information Processing Systems

\bibitem[{Zeng et~al.(2023)Zeng, Chen, Zhang, and Xu}]{zeng2023dlinear}
Zeng A, Chen M, Zhang L, et~al (2023) Are transformers effective for time series forecasting? In: Proceedings of the AAAI Conference on Artificial Intelligence, pp 11121--11128

\bibitem[{Zeng et~al.(2022)Zeng, Chu, Xu, Liu, and Quan}]{zeng2022aircraft}
Zeng W, Chu X, Xu Z, et~al (2022) Aircraft 4d trajectory prediction in civil aviation: A review. Aerospace 9(2):91

\bibitem[{Zeng et~al.(2024)Zeng, Hu, Chen, Yuan, Alam, and Xue}]{zeng2024improved}
Zeng Y, Hu M, Chen H, et~al (2024) Improved air traffic flow prediction in terminal areas using a multimodal spatial–temporal network for weather-aware ({MST-WA}) model. Advanced Engineering Informatics 62:102935

\bibitem[{Zhang et~al.(2024)Zhang, Xu, Zhang, Jiang, Alam, and Xue}]{zhang2025short}
Zhang Y, Xu S, Zhang L, et~al (2024) Short-term multi-step-ahead sector-based traffic flow prediction based on the attention-enhanced graph convolutional lstm network ({AGC-LSTM}). Neural Comput Appl 37(20):14869–14888

\bibitem[{Zhang et~al.(2023)Zhang, Guo, Zhou, Zhang, and Lin}]{zhang2023flight}
Zhang Z, Guo D, Zhou S, et~al (2023) Flight trajectory prediction enabled by time-frequency wavelet transform. Nature Communications 14:5258

\bibitem[{Zhou et~al.(2022)Zhou, Ma, Wen, Wang, Sun, and Jin}]{zhou2022fedformer}
Zhou T, Ma Z, Wen Q, et~al (2022) {FED}former: Frequency enhanced decomposed transformer for long-term series forecasting. In: Proceedings of the 39th International Conference on Machine Learning (ICML), pp 27268--27286

\end{thebibliography}

\section{Acknowledgements}
This work was supported by the related foundations. B.W. sincerely thanks Xiaofeng Gao and Jingyuan Wang for their valuable insights and discussions on air traffic flow modeling.

\section{Author contributions}
B.W., A.L., F.H., and Y.Y. conceived and led the research project. B.W. and A.L. proposed the research idea, contributed to the methodology design, and revised the manuscript. B.W., A.L., and J.Z. developed the framework, devised the neural architecture, implemented the model, and conducted the experimental studies. Y.H. contributed to the experimental evaluation and results discussion. P.H. and G.J. conducted data preprocessing, organized the dataset, and assisted with the experiments and result analysis. All authors participated in the discussion of the results. B.W., A.L., H.B., Y.Y., and Y.H.  wrote the manuscript with input from all authors. F.H., Y.Y., Y.H., and T.L. supervised and sponsored the study. B.W., A.L., F.H., Y.Y., and Y.H. approved the submission and accepted responsibility for the overall integrity of the paper.

\section{Competing interests}
The authors declare no competing interests.

\section{Additional information}
Correspondence and requests for materials should be addressed to Feng Hong.




\end{document}